\documentclass[10pt,twocolumn,letterpaper]{article}

\usepackage{cvpr}              

\usepackage[most]{tcolorbox}
\usepackage{booktabs}
\usepackage{multirow}
\definecolor{cvprblue}{rgb}{0.21,0.49,0.74}
\usepackage[pagebackref,breaklinks,colorlinks,allcolors=cvprblue]{hyperref}

\def\paperID{1901} 
\def\confName{CVPR}
\def\confYear{2026}
\definecolor{darkgray}{RGB}{169,169,169} 
\definecolor{lightgray}{RGB}{230,230,230} 
\title{OddGridBench: Exposing the Lack of Fine-Grained Visual Discrepancy Sensitivity in Multimodal Large Language Models}

\author{
Tengjin Weng\textsuperscript{1,2},
Wenhao Jiang\textsuperscript{2,}\thanks{Corresponding author.},
Jingyi Wang\textsuperscript{4}, 
Ming Li\textsuperscript{2},
Lin Ma\textsuperscript{5},
Zhong Ming\textsuperscript{1,2,3,*} \\
\textsuperscript{1}College of Computer Science and Software Engineering, Shenzhen University\\  
\textsuperscript{2}Guangdong Laboratory of Artificial Intelligence and Digital Economy (SZ) \\
\textsuperscript{3}Shenzhen Technology University\\
\textsuperscript{4}Tsinghua Shenzhen International Graduate School \\
\textsuperscript{5}Meituan \\
}

\begin{document}
\maketitle
\begin{abstract}
Multimodal large language models (MLLMs) have achieved remarkable performance across a wide range of vision language tasks. However, their ability in low-level visual perception, particularly in detecting fine-grained visual discrepancies, remains underexplored and lacks systematic analysis.
In this work, we introduce OddGridBench, a controllable benchmark for evaluating the visual discrepancy sensitivity of MLLMs. OddGridBench comprises over 1,400 grid-based images, where a single element differs from all others by one or multiple visual attributes such as color, size, rotation, or position.
Experiments reveal that all evaluated MLLMs, including open-source families such as Qwen3-VL and InternVL3.5, and proprietary systems like Gemini-2.5-Pro and GPT-5, perform far below human levels in visual discrepancy detection.
We further propose OddGrid-GRPO, a reinforcement learning framework that integrates curriculum learning and distance-aware reward. By progressively controlling the difficulty of training samples and incorporating spatial proximity constraints into the reward design, OddGrid-GRPO significantly enhances the model's fine-grained visual discrimination ability.
We hope OddGridBench and OddGrid-GRPO will lay the groundwork for advancing perceptual grounding and visual discrepancy sensitivity in multimodal intelligence.
Code and dataset are available at \href{https://wwwtttjjj.github.io/OddGridBench/}{https://wwwtttjjj.github.io/OddGridBench/}.
\end{abstract}

\vspace{-2mm}
\section{Introduction}
\label{sec:intro}

\begin{figure}[!t]
\centering
\includegraphics[width=\linewidth]{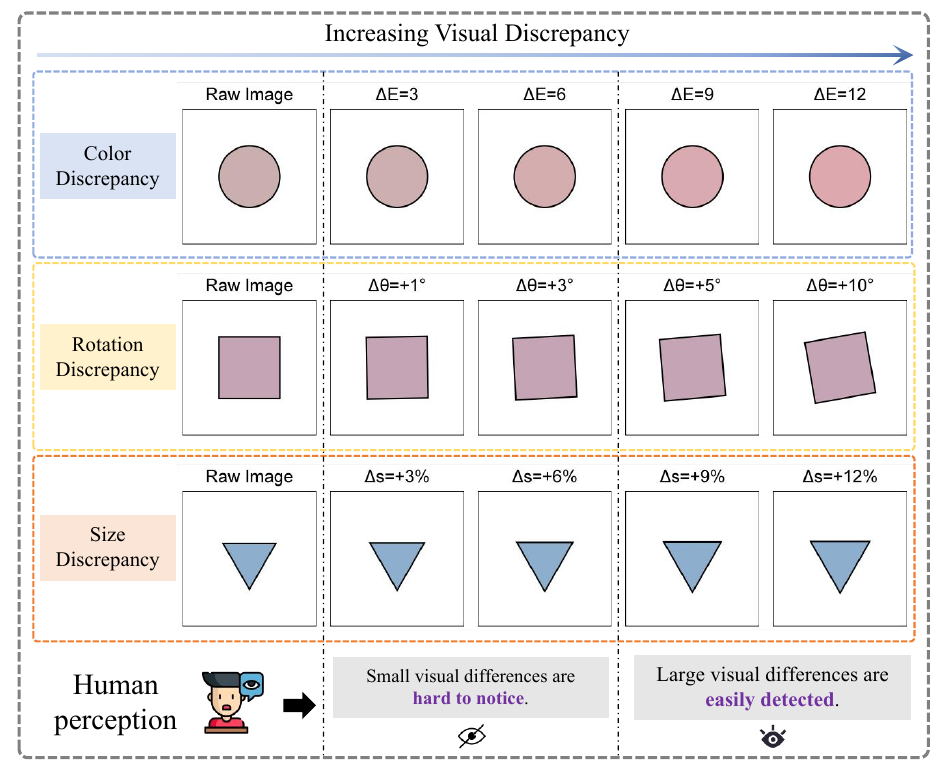}
\caption{Illustration of human perceptual visual discrepancy sensitivity, showing the transition from imperceptible to perceptible visual differences in color, rotation, and size.
}
\label{fig:JND}
\vspace{-3mm}
\end{figure}

\begin{figure*}[!t]
\centering
\includegraphics[width=\linewidth]{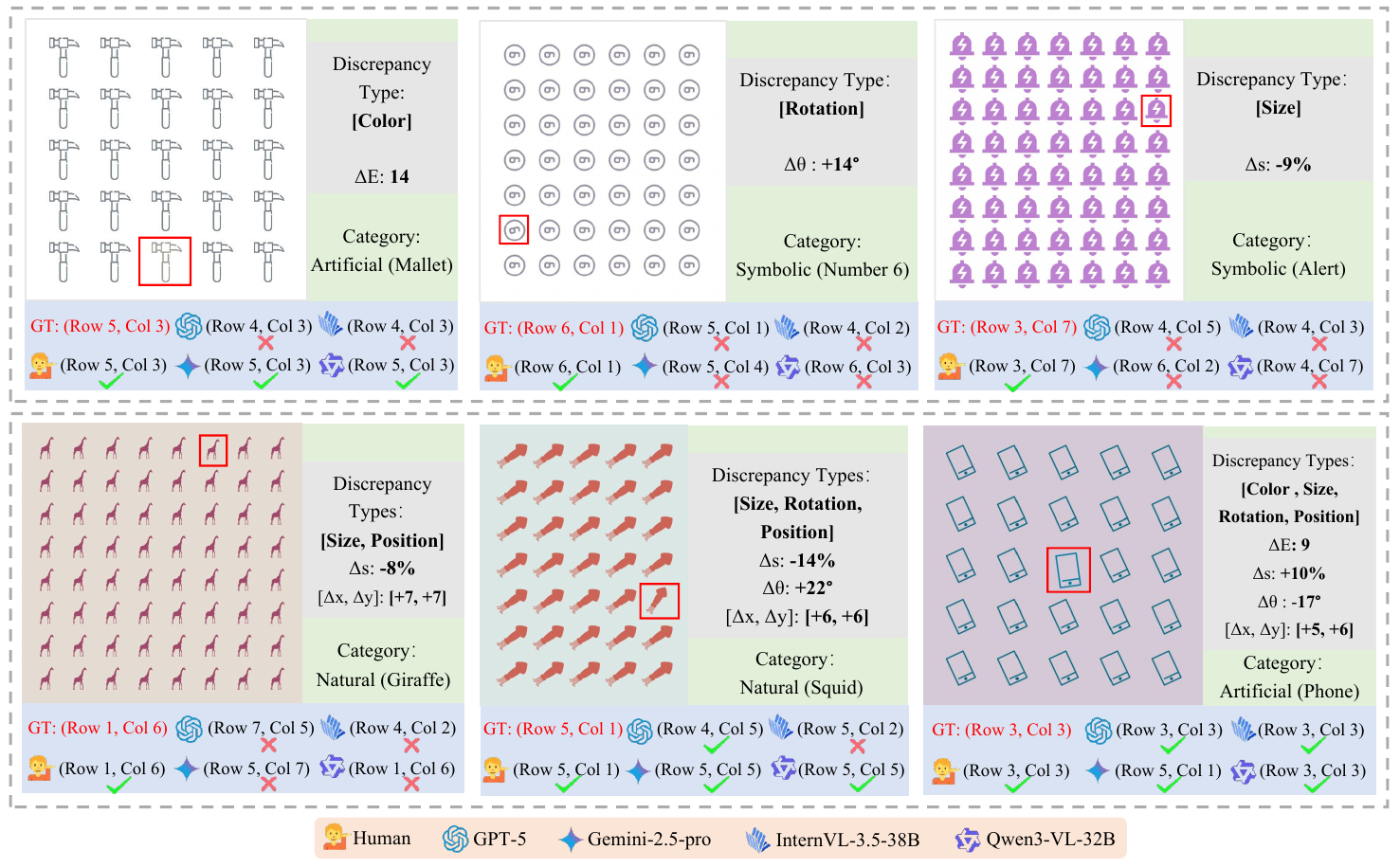}
\caption {Overview of OddGridBench. OddGridBench encompasses four primary visual attributes, including color, size, rotation, and position, and supports both single-attribute and multi-attribute discrepancy compositions, providing a systematic framework for evaluating the perceptual discrepancy sensitivity of MLLMs.}
\label{fig:Oddgird}
\end{figure*}

Human vision exhibits remarkable sensitivity to subtle visual discrepancies.
This phenomenon is described within the frameworks of just noticeable difference and the pop-out effect~\cite{weber1834pulsu,wolfe1994guided}, suggesting that the visual system can rapidly identify any element that disrupts the uniformity of a visual field.
As shown in Figure~\ref{fig:JND}, gradual changes in low-level visual attributes such as color, orientation, and size can trigger a perceptual transition from indistinguishable to salient stimuli. 
These observations highlight the human visual system’s inherent ability to perceive fine-grained variations even in cluttered or homogeneous scenes.

Multimodal Large Language Models (MLLMs) have achieved remarkable progress in high-level reasoning and multimodal understanding~\cite{alayrac2022flamingo,chen2024far,li2023blip,liu2024improved, tan2025wmarkgpt}.
Existing benchmarks for MLLMs mainly focus on high-level visual understanding in natural scenes, such as image captioning, referring expression comprehension, and visual commonsense reasoning~\cite{yang2025embodiedbench, xu2025visulogic, chen2025co}.
Beyond general perception, other benchmarks emphasize symbolic reasoning and semantic interpretation, including mathematical problem solving, diagram understanding, and structured visual question answering~\cite{dahlgren2022clevr,lu2023mathvista,kamoi2024visonlyqa,hu2025emobench,weng2025visnumbench}.
However, existing evaluations of MLLMs often overlook an essential aspect of human vision, the ability to detect fine-grained discrepancies in visual scenes. \textbf{Such perceptual sensitivity is a prerequisite for robust spatial reasoning, object understanding, grounding, and visual question answering, and weakness at this foundational layer undermines the reliability of higher-level MLLMs' capabilities.}

\begin{figure}[!t]
\centering
\includegraphics[width=\linewidth]{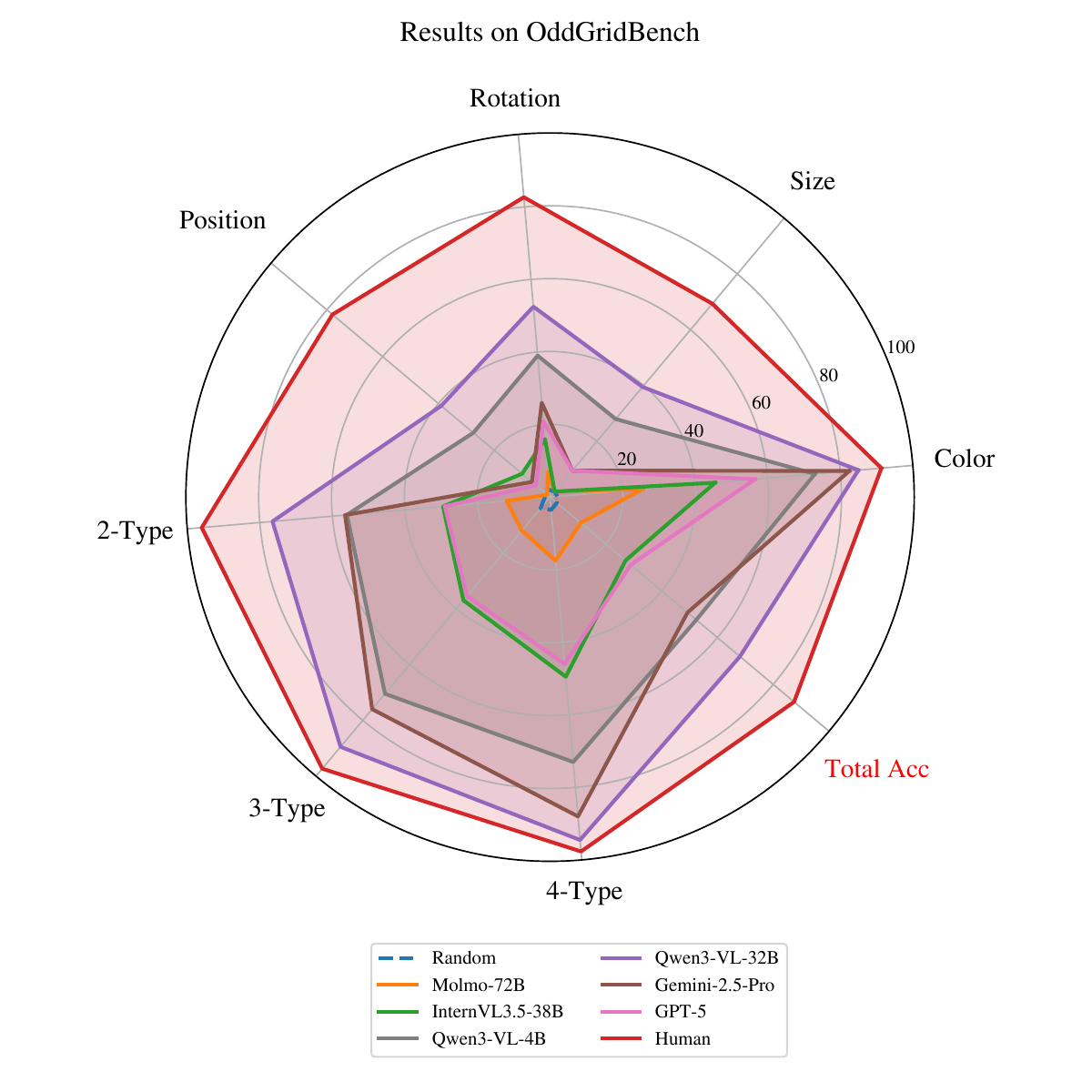}
\caption{Evaluation results of MLLMs on the OddGridBench. Human performance significantly surpasses all evaluated MLLMs across color, size, rotation, and position dimensions, as well as multi-type combinations.
}
\vspace{-2mm}
\label{fig:radar}
\end{figure}

In this work, we introduce OddGridBench, a controllable and scalable benchmark based on the Odd-One-Out paradigm~\cite{treisman1980feature}, designed to evaluate the visual perceptual discrepancy sensitivity of MLLMs systematically.
As illustrated in Figure~\ref{fig:Oddgird}, each instance presents a grid of visually similar icons, where one element subtly differs from the others in a single or multiple low-level visual attributes such as color ($\Delta E$), size ($\Delta s$), rotation ($\Delta \theta$), or position ($[\Delta x, \Delta y]$).
By isolating perceptual differences from high-level semantics, OddGridBench allows a direct evaluation of MLLMs’ ability to perceive visual discrepancies.
We evaluated 19 MLLMs on OddGridBench and found that even state-of-the-art models perform poorly in fine-grained visual discrepancy perception, as shown in Figure~\ref{fig:radar}.
We further propose OddGrid-GRPO, a reinforcement learning framework designed to enhance MLLMs’ perceptual sensitivity through curriculum learning and spatially guided reward.
The main contributions are summarized as follows:
\vspace{-2mm}
\begin{enumerate}
    \item We present OddGridBench, a scalable, controllable benchmark for evaluating the perceptual discrepancy capabilities of MLLMs.
    By generating grid-based images in a parameterized space that continuously controls color, scale, rotation, and position, OddGridBench enables quantitative and systematic analysis of model sensitivity across multiple perceptual dimensions.
   
    
    \item We conduct comprehensive experiments on a broad set of state-of-the-art 
    open-source and proprietary MLLMs, revealing consistent and previously underexplored 
    failure patterns in fine-grained perceptual discrimination across all model families.

    \item We propose OddGrid-GRPO, by combining curriculum learning with a distance-aware reward that provides continuous perceptual feedback, the framework progressively enhances discrepancy sensitivity and achieves more fine-grained visual discrimination.
\end{enumerate}

\vspace{-2mm}
\section{Related Works}
\label{sec:relate}
\subsection{MLLMs and Benchmarks}

Recent advances in MLLMs have demonstrated remarkable capabilities across diverse visual and linguistic tasks, driven by large-scale pre-training on paired image-text data. Both open-source~\cite{yin2025sail,an2025llava,wang2025internvl3,abdin2024phi,deitke2025molmo,chen2025januspro,chen2024expanding,dai2024nvlm} and proprietary systems~\cite{openai2023gpt4v,pichai2024our,anthropic2024claude} have achieved impressive performance in domains such as mathematical reasoning~\cite{zhang2024mavis}, chart understanding~\cite{masry2023unichart}, and medical image interpretation~\cite{li2024llava}. 
Such rapid progress has spurred the development of numerous benchmarks for evaluating multimodal understanding and reasoning~\cite{weng2025visnumbench,fu2025video,jiang2025mac,ying2024mmt,yang2024thinking, zhao2025favchat}. 
However, existing benchmarks for MLLMs mainly focus on high-level semantic reasoning, while overlooking fine-grained low-level visual perception, which serves as the foundation for accurate understanding and reasoning.

\subsection{Fine-Grained Visual Discrepancy Sensitivity}
Fine-grained visual discrepancy sensitivity represents a fundamental aspect of low-level visual perception, reflecting a model’s ability to distinguish subtle differences among visually similar elements. 
The natural starting point for investigating perceptual sensitivity is the Odd-One-Out paradigm, which has long been used in cognitive psychology and vision science to study perceptual discrimination and attentional mechanisms. 
In visual recognition and representation learning, Odd-One-Out or anomaly detection setups have been employed to evaluate a model’s ability to capture perceptual distinctiveness and local irregularities~\cite{bhunia2025odd, born2024evaluating, akbarinia2025exploring}. 
Bhunia \textit{et al.}~\cite{bhunia2025odd} proposed an Odd-One-Out detection framework for visual anomaly recognition, while Akbarinia ~\cite{akbarinia2025exploring} investigated visual sensitivity in neural architectures through psychophysical-style stimuli. 
These studies demonstrate that machine perception can, to some extent, replicate human-like discrimination behavior when exposed to controlled visual perturbations. 
However, most existing implementations rely on small-scale datasets or handcrafted stimuli, which limit their generalizability across diverse perceptual dimensions.
Moreover, prior Odd-One-Out or anomaly-detection frameworks are tailored to visual encoders or recognition networks trained with dense pixel-level supervision, making them unsuitable for MLLMs whose architectures do not inherently support region-level perception or localization.
These limitations highlight the necessity of establishing an evaluation framework with dynamic and interpretable feedback to better understand how models perceive visual perturbations.

\vspace{-1mm}
\subsection{Reinforcement Learning for Visual Alignment}
Reinforcement learning (RL) enhances visual perception and alignment in multimodal models, enabling them to learn from feedback-driven supervision rather than static labels.
It has become a central paradigm for aligning large language and multimodal models with human preferences~\cite{zhou2023lima,chaudhari2025rlhf}.
Recent progress in RL for model alignment has been driven by the Group Relative Policy Optimization (GRPO) algorithm, the core optimization paradigm behind DeepSeek-V3~\cite{liu2024deepseek}.
Building on this foundation, recent studies~\cite{huang2025high,bai2025univg,cao2025ground} have extended GRPO-based training beyond text-only alignment to multimodal domains, where RL enhances image reasoning and perceptual grounding.
These developments highlight a promising direction for aligning large multimodal models both semantically and perceptually, achieving finer-grained consistency between visual inputs and model reasoning.
However, existing RL-based alignment approaches primarily focus on outcome-level rewards capturing semantic correctness or linguistic coherence, while overlooking perceptual alignment between model predictions and human visual judgments.
Developing effective reinforcement strategies for fine-grained visual discrepancy perception requires a paradigm that operates directly at the perceptual level, enabling models to discriminate subtle visual variations independent of semantic supervision.

\begin{figure*}[!t]
\centering
\includegraphics[width=\linewidth]{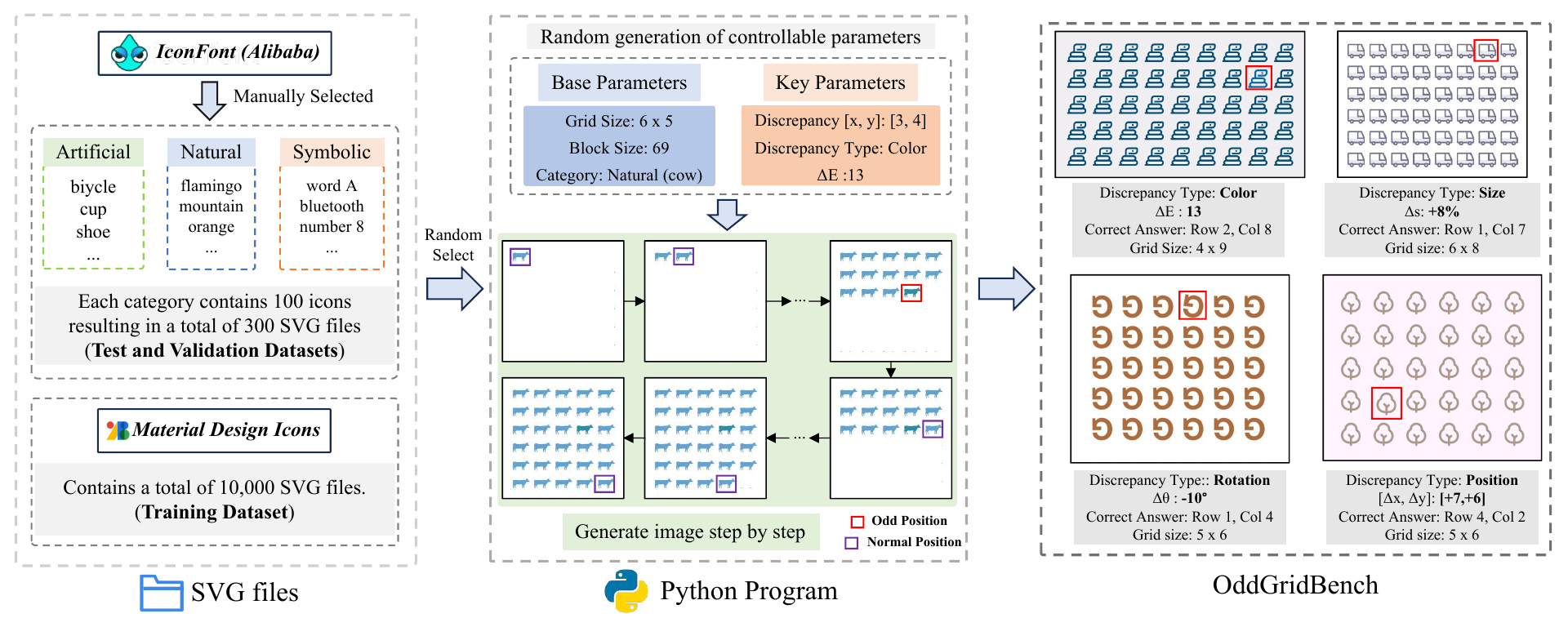}
\caption{Overview of the OddGridBench data generation pipeline, which constructs grid-based images from collected icons under precisely controlled perceptual conditions for evaluating visual discrepancy sensitivity.
}
\label{fig:data_generation}
\end{figure*}
\section{OddGrid Benchmark}
\label{sec:method}

\subsection{Overview}
OddGridBench is designed to evaluate model sensitivity to fine-grained visual discrepancies under controlled conditions. 
As shown in Figure~\ref{fig:data_generation}, icons are collected from a large-scale vector repository and synthesized into grid images within a rigorously parameterized space, with all generation metadata recorded for full reproducibility and quantitative analysis. Synthetic icons are used intentionally to provide precise, 
psychophysics-style control over perceptual variables that real-world images cannot offer.
OddGridBench contains \textbf{1,400 test samples} spanning seven types: four single attributes (Color, Size, Rotation, Position) and three multi-attribute combinations (2-Type, 3-Type, and 4-Type), with 200 samples per attribute.
An additional \textbf{400 validation} images and \textbf{30,000 training images} are generated under the same conditions to ensure consistency across splits. 
\vspace{-2mm}
\subsection{Icon Source}
\label{sec:method:source}

We collect icons from two sources: IconFont~\cite{iconfont} and Material Design Icons~\cite{material-icons}.
All icons are standardized into scalable vector graphics (SVG) format to ensure resolution independence and visual consistency.
For the {test} and {validation} datasets, we manually select 300 SVG icons from IconFont and categorize them into three semantic groups.
\begin{itemize}
    \item \textbf{Artificial}: Man-made objects such as bicycles, cups, and shoes.
    \item \textbf{Natural}:  Natural elements like flamingos, mountains, and oranges.
    \item \textbf{Symbolic}:  Abstract or textual symbols such as letters, numbers, and logos.
\end{itemize}
Each group contains 100 distinct icons, resulting in 300 manually curated SVG files in total.  
For the {training dataset}, we collect 10,000 SVG icons from {Material Design Icons}, which provide diverse yet standardized vector representations.

\begin{figure*}[!t]
\centering

\includegraphics[width=\linewidth]{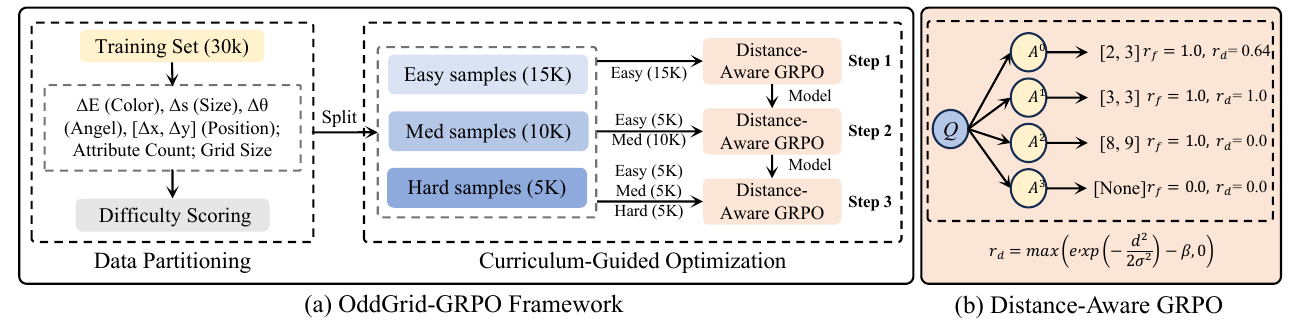}
\caption{Overview of OddGrid-GRPO framework. OddGrid-GRPO integrates curriculum-guided optimization with spatially guided reward shaping to enhance perceptual grounding and improve fine-grained visual discrimination in MLLMs.}
\label{fig:oddrl}
\end{figure*}

\subsection{Controlled Image Generation}
\label{sec:method:generation}
Each OddGridBench image is generated using a parameterized Python program for precise control of layout, attributes, and perceptual differences.

\noindent \textbf{Basic image layout parameters.}  
The spatial structure of each image is primarily defined by two parameters:

\begin{itemize}
    \item \textbf{Grid size:}  
    Each image adopts a grid layout consisting of 5 to 9 rows and columns.  
    Varying the grid size controls the spatial density of items and the overall perceptual complexity of the scene.  

    \item \textbf{Block size:}  
    Each icon is rendered within a square region whose size is randomly sampled between 60 and 80 pixels.    
    Randomized block sizes are used within a predefined range to introduce moderate variation while keeping the layout visually consistent.
\end{itemize}

\vspace{-1mm}
\noindent \textbf{Visual discrepancy parameters.}
The perceptual variation between the odd item and the distractors is controlled by four types of parameters, corresponding to distinct perceptual attributes:
\begin{itemize}
    \item \textbf{Color discrepancy ($\Delta E$):} 
    The color contrast is defined in the CIE–Lab color space. 
    The base and odd items are assigned Lab values, and their perceptual distance $\Delta E$ is sampled from a predefined range ([5, 20]), ensuring controlled transitions from imperceptible to clearly perceptible differences.

    \item \textbf{Size discrepancy ($\Delta s$):}  
    The odd item is generated by slightly resizing the base icon, either reducing it to 85\%–95\% or enlarging it to 105\%–115\% of its original size.

    \item \textbf{Rotation discrepancy ($\Delta \theta$):} 
    The odd item is generated by rotating the base icon clockwise or counterclockwise within a controlled angular range, typically between $-25^\circ$ and $-5^\circ$ or between $5^\circ$ and $25^\circ$.
    
    \item \textbf{Position discrepancy ($[\Delta x, \Delta y]$):}  
    The odd item is shifted slightly away from its grid center in both horizontal and vertical directions, typically by 5\%–12\% of the block size.  
    This small displacement produces a subtle but clearly localized positional deviation, simulating mild spatial misalignment.
    
\end{itemize}
In addition to single-type discrepancies, OddGridBench also includes multi-type combinations, 
where the odd item simultaneously differs from the base items in multiple perceptual dimensions. 
For example, a 2-Type case may jointly vary color and size, while a 4-Type case integrates differences in color, size, rotation, and position. 
These mixed conditions increase task complexity by introducing potential interactions among visual attributes.

\section{OddGrid-GRPO}
\label{sec:method}

\begin{table*}[!t]
\centering
    \renewcommand\arraystretch{1.0}
    \caption{
    Accuracies (\%) of various MLLMs on the OddGridBench dataset. 
    The benchmark evaluates fine-grained visual discrimination across four perceptual dimensions as well as their multi-type combinations. 
    \colorbox{darkgray}{Dark gray} and \colorbox{lightgray}{Light gray} indicate the best and second-best results among all models, respectively. 
    }
    \label{tabel_main}
    \large
\scalebox{0.85}{
\begin{tabular}{ccccccccc}
\toprule[1pt]
Method                 & Color & Size  & Rotation & Position & 2-Type & 3-Type & 4-Type & Total \\ \midrule
Random                 & 2.00  & 2.00  & 2.00    & 2.00     & 1.50   & 4.00   & 3.50   & 2.43  \\ \midrule
\multicolumn{9}{c}{Open-source MLLMs}                                                          \\ \midrule
Phi-3.5-vision         & 2.00  & 2.00  & 2.00    & 1.50     & 1.50   & 1.50   & 2.00   & 1.79  \\
SAIL-VL2-2B            & 13.50 & 2.50  & 3.00    & 2.00     & 7.50   & 5.00   & 9.50   & 6.14  \\
SAIL-VL2-8B            & 45.00 & 2.50  & 5.00    & 2.50     & 21.50  & 24.00  & 30.50  & 18.71 \\
LLaVA-OneVision-1.5-4B & 27.50 & 3.00  & 7.50    & 1.00     & 14.00  & 17.50  & 26.00  & 13.79 \\
LLaVA-OneVision-1.5-8B & 39.50 & 7.50  & 9.00    & 4.50     & 21.00  & 37.00  & 47.50  & 23.71 \\
LLaVA-v1.6-34B         & 3.50  & 2.00  & 2.50    & 2.50     & 3.00   & 3.00   & 2.50   & 2.71  \\
InternVL3.5-38B        & 45.50 & 2.00  & 16.00   & 10.00    & 29.50  & 37.00  & 49.50  & 27.07 \\
Molmo-72B              & 25.50 & 2.00  & 7.00    & 1.00     & 12.00  & 12.00  & 17.50  & 11.00 \\
Qwen2.5-VL-7B          & 2.00  & 1.50  & 0.50    & 0.00     & 2.50   & 2.50   & 5.00   & 2.00  \\
Qwen2.5-VL-72B         & 25.00 & 1.00  & 5.00    & 2.50     & 10.50  & 17.50  & 17.50  & 11.29 \\
Qwen3-VL-2B            &23.00  &5.00   &12.50    &7.00     &19.00  &22.50  &31.00   & 17.14 \\
Qwen3-VL-4B            & {73.00} & \colorbox{lightgray}{28.00} & \colorbox{lightgray}{39.00}   & \colorbox{lightgray}{27.50}    & {56.00}  & {70.50}  & {73.00}  & \colorbox{lightgray}{52.43} \\
Qwen3-VL-8B            & 72.00 & 20.50 & {35.50}   & 19.00    & 55.50  & 67.00  & 71.50  & 48.71 \\
Qwen3-VL-30B (Moe)           & 62.50 & {21.00} & 20.50   & {25.00}    & 47.00  & 52.50  & 62.00  & 41.50 \\ 
Qwen3-VL-32B           &\colorbox{darkgray}{85.00} &\colorbox{darkgray}{39.50} &\colorbox{darkgray}{52.50} &\colorbox{darkgray}{39.00} &\colorbox{darkgray}{76.50} &\colorbox{darkgray}{89.50} &\colorbox{darkgray}{94.50} &\colorbox{darkgray}{68.07} \\
\midrule
\multicolumn{9}{c}{Proprietary MLLMs}                                                          \\ \midrule
Gemini-2.0-Flash       & 53.50 & 5.00  & 16.00   & 5.00     & 32.50  & 43.50  & 47.50  & 29.00 \\
Gemini-2.5-Flash       & 70.00 & 7.00  & 21.00   & 5.00     & 48.50  & 54.50  & 72.50  & 39.79 \\
Gemini-2.5-Pro         & \colorbox{lightgray}{82.50} & 9.50  & 26.00   & 6.50     & \colorbox{lightgray}{56.50}  & \colorbox{lightgray}{76.00}  & \colorbox{lightgray}{88.00}  & {49.29} \\
GPT-5                  & 56.50 & 9.50  & 21.00   & 5.00     & 29.00  & 35.50  & 46.00  & 28.93 \\ \midrule
Human                  & 91.33 & 69.33 & 82.67   & 78.00    & 96.00  & 97.33  & 97.67  & 87.47 \\  \midrule
\end{tabular}}
\end{table*}

\subsection{Motivation and Overview}
Current MLLMs exhibit low sensitivity to fine-grained visual discrepancies and often fail to reason over structured spatial layouts, frequently miscounting or confusing grid indices even when the anomalous region is visually clear.
To overcome these limitations, we propose OddGrid-GRPO, as illustrated in Fig.~\ref{fig:oddrl}, which integrates two components:
\begin{itemize}
    \item \textbf{Curriculum-guided optimization.} 
    Training begins with coarse visual differences and gradually shifts toward subtle ones, which stabilizes the RL process and enables the model to acquire fine-grained perceptual sensitivity in a more human-like manner.
    \item  \textbf{Distance-aware reward formulation.} We design a distance-based reward that provides continuous feedback proportional to the spatial proximity between predicted and ground-truth coordinates, offering richer supervision than the binary signal in standard GRPO and enabling better modeling of spatial dependencies.
    
\end{itemize}
More details of the curriculum-guided optimization~\ref{sec:curriculum} and distance-aware reward formulation~\ref{sec:reward} are provided in the following subsections.

\subsection{Curriculum-Guided Optimization}
\label{sec:curriculum}
Each sample is assigned a continuous difficulty score determined by three key factors: grid size, odd item attribute count, and fine-grained perturbation parameters.
Larger grid sizes, fewer odd item attribute counts, and smaller perturbation magnitudes correspond to higher difficulty scores, indicating that such samples are harder to distinguish.
\vspace{-2mm}
\paragraph{Data Partitioning.} 
Samples are sorted according to their difficulty scores and divided into three subsets with a ratio of {Easy} (15K), {Medium} (10K), and {Hard} (5K), corresponding to progressively increasing perceptual difficulty. The {Easy} subset mainly contains grids with large, salient discrepancies, while the {Hard} subset includes subtle, near-threshold variations that require fine-grained perceptual sensitivity. Intermediate samples form the {Medium} subset, bridging the two extremes.
\vspace{-2mm}
\paragraph{Progressive Training Schedule.} 
The model is optimized progressively in three stages. 
In {Step 1}, training begins with 15K easy samples, allowing the policy to establish coarse spatial alignment. 
In {Step 2}, the model is trained on a mixed set of 5K easy and 10K medium samples to improve generalization under moderate visual discrepancies. 
In {Step 3}, 5K hard samples are introduced along with a subset of easier ones to stabilize optimization and strengthen fine-grained perceptual discrimination. 
Such progressive learning from easy to hard conditions prevents premature convergence and encourages refined visual sensitivity.

\subsection{Distance-Aware Reward Formulation}
\label{sec:reward}

GRPO employs a binary correctness signal, where the model receives a reward of~1 for an exact match and~0 otherwise. 
While this formulation is suitable for discrete reasoning tasks, it is suboptimal for perceptual localization, where predictions that are spatially close to the correct cell should not be penalized as harshly as entirely incorrect ones. 
To provide a smoother and more informative learning signal, we design a distance-aware reward that continuously scales with the spatial proximity between the predicted and ground-truth positions.

For each prediction, we compute the Euclidean distance between the predicted and ground-truth grid positions as
\[
d = \sqrt{(r_p - r_g)^2 + (c_p - c_g)^2},
\]
where $(r_p, c_p)$ and $(r_g, c_g)$ denote the row and column indices of the predicted and ground-truth cells, respectively. 
If the prediction exactly matches the ground truth, a perfect reward of $r_{\text{d}} = 1$ is assigned. 
Otherwise, the reward decays smoothly with increasing spatial distance according to
\[
r_{\text{d}} = \max \!\left( \exp\!\left(-\frac{d^2}{2\sigma^2}\right) - \beta , \, 0 \right),
\]
where $\sigma$ is a scale-adaptive parameter proportional to the grid diagonal 
($\sigma = \lambda \sqrt{n_\text{rows}^2 + n_\text{cols}^2}$), 
and $\beta$ is a small bias term that suppresses rewards for distant predictions. 
In practice, $\lambda$ is set to~0.25 and $\beta$ to~0.3, controlling the smoothness of the Gaussian decay and the cutoff threshold for low-confidence samples.

We retain a small format reward to ensure structural consistency in model outputs, and define the overall optimization reward as
\[
r_{\text{overall}} = (1 - \omega) \, r_{\text{d}} + \omega \, r_{\text{f}},
\]
where $\omega$ denotes the format weight and is empirically set to~0.2. 
OddGrid-GRPO provides a principled way to align reinforcement signals with geometric proximity, effectively enhancing row–column reasoning and spatial alignment.

\section{Experiments}
\label{sec:experiment}

\begin{figure*}[!t]
\centering
\includegraphics[width=\linewidth]{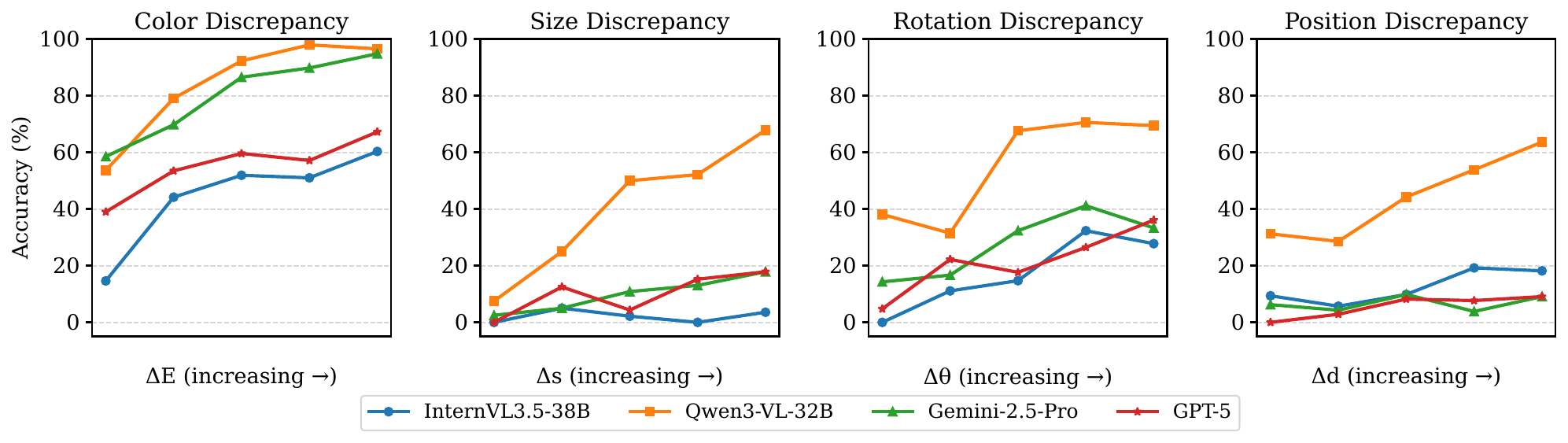}
\caption{
Accuracy across varying perceptual discrepancy magnitudes in four visual dimensions.
The x-axis represents the incremental difference in each attribute 
(\(\Delta d = \sqrt{(\Delta x)^2 + (\Delta y)^2}\)). 
As the discrepancy increases from imperceptible to clearly perceptible levels, model performance gradually improves.
}
\label{fig:line}
\end{figure*}

\begin{table}[!t]
\centering
\renewcommand\arraystretch{1.0}
\caption{Comparison of strict accuracy (Acc), Tolerance Accuracy (TolAcc), and Labeled Grid Accuracy (LabeledAcc). 
Values in green indicate improvements relative to strict accuracy.}
\label{table_ablation1}
\large
\scalebox{0.82}{
\begin{tabular}{cccc}
\toprule[1pt]
Method          & Acc   & TolAcc                                               & LabeledAcc                                              \\ \midrule
InternVL3.5-38B & 27.07 & 53.29\scriptsize\textcolor{green}{\((\uparrow +26.22)\)}  & 56.40\scriptsize\textcolor{green}{\((\uparrow +19.33)\)} \\
Qwen3-VL-4B     & 52.43 & 74.14\scriptsize\textcolor{green}{\((\uparrow +21.71)\)}  & 70.40\scriptsize\textcolor{green}{\((\uparrow +17.97)\)} \\
Qwen3-VL-32B & 68.07 & 75.43\scriptsize\textcolor{green}{\((\uparrow +7.36)\)} & 73.80\scriptsize\textcolor{green}{\((\uparrow +5.73)\)} \\
Gemini-2.5-Pro  & 49.29 & 61.21\scriptsize\textcolor{green}{\((\uparrow +11.92)\)}  &
58.40\scriptsize\textcolor{green}{\((\uparrow +9.11)\)}  \\ \midrule
\end{tabular}}
\end{table}

\begin{table*}[!t]
\centering
\renewcommand\arraystretch{1.0}
\caption{
Quantitative comparison of OddGrid-GRPO with existing reinforcement learning methods (upper block) 
and its ablated variants (lower block) across all perceptual attribute types on {OddGridBench}.
}
\label{table_method}
\large
\scalebox{0.8}{
\begin{tabular}{ccccccccc}
\toprule[1pt]
Method                    & Color          & Size           & Rotation        & Position       & 2-Type         & 3-Type         & 4-Type         & Total          \\ \midrule
Baseline                      & 23.00          & 5.00           & 12.50          & 7.00           & 19.00          & 22.50          & 31.00          & 17.14          \\
GRPO                      & 88.50          & 44.00          & 67.50          & 41.50          & 78.50          & 83.00          & 93.00          & 70.86          \\ 
GSPO                      & 70.00          & 55.00          & 81.50          & 59.00          & 85.50          & 85.50         & 95.00         & 75.93          \\ \midrule
OddGrid-GRPO (w/o $r_{\text{d}}$)   & 87.50          & 44.50          & 67.00          & 45.50          & 80.50          & 91.00          & 91.50          & 72.50          \\
OddGrid-GRPO (w/o Cur-Guided) & 87.50          & 60.50          & 69.00          & 64.00          & 84.00          & 88.50          & 95.50          & 78.43          \\
OddGrid-GRPO              & \textbf{89.50} & \textbf{64.50} & \textbf{80.50} & \textbf{64.50} & \textbf{90.50} & \textbf{91.50} & \textbf{97.50} & \textbf{82.64}    \\ \midrule
\end{tabular}}
\end{table*}

\paragraph{Evaluation Settings}
We evaluate 19 representative MLLMs from 9 model families, covering both open-source and proprietary systems.
The open-source models include Phi-3.5-Vision~\cite{abdin2024phi}, SAIL-VL2 (2B, 8B)~\cite{yin2025sail}, LLaVA-OneVision-1.5 (4B, 8B)~\cite{an2025llava}, LLaVA-v1.6-34B~\cite{liu2023llava}, InternVL3.5-38B~\cite{wang2025internvl3}, Molmo-72B~\cite{deitke2025molmo}, and Qwen-VL series including Qwen2.5-VL (7B, 72B)~\cite{qwen2.5-VL} and Qwen3-VL (2B, 4B, 8B, 30B, 32B)~\cite{qwen3technicalreport}.
The proprietary models consist of Gemini-2.0-Flash, Gemini-2.5-Flash, Gemini-2.5-Pro~\cite{comanici2025gemini}, and GPT-5~\cite{hurst2024gpt}.
For human evaluation, we randomly sample 350 instances (50 from each perceptual attribution). Each sample is independently annotated by three trained human evaluators who identify the odd element based solely on visual cues. Final human accuracy is computed as the average correctness across annotators. More details can be found in Appendix~\ref{exp}.
\vspace{-2mm}
\paragraph{Training Settings.}
We adopt Qwen3-VL-2B as the base model and train it under the {EasyR1}~\cite{zheng2025easyr1} RL framework.
The optimizer is {AdamW} with a learning rate of $1\times10^{-6}$, weight decay of $1\times10^{-2}$, gradient clipping at 1.0, and a KL-penalty coefficient of $1\times10^{-2}$.
Each rollout generates {three samples}, and training is performed for 100 steps on 4×A800 GPUs with a global batch size of 256.
All experiments use identical configurations for fair comparison.

\subsection{Evaluation Results on OddGridBench}
Table~\ref{tabel_main} summarizes the quantitative results across all evaluated models.
Models like Phi-3.5-Vision and LLaVA-v1.6-34B perform near random, while SAIL-VL2-8B and LLaVA-OneVision-8B show slight improvements in color and multi-type tasks but still struggle with rotation and position.
Qwen3-VL-32B achieves the highest overall accuracy (68.07\%) across all models.
Notably, even Qwen3-VL-2B and Qwen3-VL-4B outperform much larger models such as InternVL3.5-38B and Molmo-72B, demonstrating that fine-grained perceptual discrimination depends more on data alignment and perceptual coupling than on parameter scale alone. 
For proprietary systems, performance does not show a clear advantage over leading open-source models. Gemini-2.5-Pro achieves the best overall score (49.29\%) among proprietary systems, excelling in color and multi-type categories but still lagging behind Qwen3-VL-4B. 
Gemini-2.0-Flash, Gemini-2.5-Flash, and GPT-5 reach comparable levels to mid-scale open-source models.
Existing MLLMs exhibit limited sensitivity to subtle visual discrepancies, particularly under rotation and position perturbations. By contrast, human participants achieve a total accuracy of 87.47\%, maintaining stable performance across all categories and revealing a pronounced gap between human and model-level perception. 


\subsection{Further Analysis}

\noindent \textbf{Sensitivity to Perceptual Discrepancy Magnitude.}
\label{Sensitivity}
To investigate the effect of attribute discrepancy magnitude on model performance, we analyze MLLMs' accuracy under incremental variations of a single attribute, as shown in Figure~\ref{fig:line}.
For each perceptual attribute, the full range of discrepancy values is divided into five equal intervals, corresponding to a progressive increase in perceptual difficulty from nearly imperceptible to clearly distinguishable differences. As the discrepancy level increases, model accuracy gradually improves. 
Color discrepancy yields the most significant performance gains, whereas geometric attributes such as Rotation and Position exhibit much slower improvement. 
This trend indicates that current MLLMs rely heavily on coarse visual differences while lacking fine-grained perceptual sensitivity.

\begin{figure}[!t]
\centering
\includegraphics[width=\linewidth]{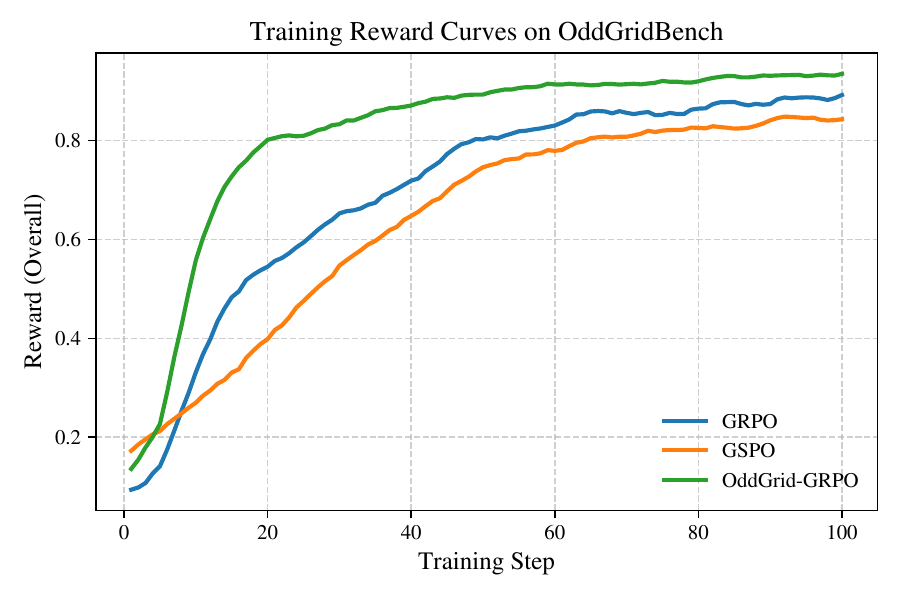}
\caption{Training curves of GRPO, GSPO, and OddGrid-GRPO. OddGrid-GRPO converges faster and achieves higher final rewards, indicating more stable and efficient optimization.}
\vspace{-2mm}
\label{fig:reward_curve}
\end{figure}

\noindent \textbf{Analysis of Localization Accuracy.}
To further examine the nature of model errors, we introduce two extra metrics for accuracy: 
(1) Tolerance Accuracy (TolAcc): counts a prediction as correct if it lies within one row and column of the ground-truth cell (\(|\Delta \text{row}| \le 1, |\Delta \text{col}| \le 1\)), capturing coarse spatial sensitivity. 
(2) Labeled Grid Accuracy (LabeledAcc): re-evaluates model performance when the input grid image explicitly displays row and column indices, allowing the model to make predictions based on visible coordinate information.
As shown in Table~\ref{table_ablation1}, all evaluated models achieve clear improvements under the relaxed metrics. 
For example, Qwen3-VL-4B increases from 52.43\% to 74.14\% in TolAcc, suggesting that most of its predictions fall near the correct grid but lack precise localization. 
Similar patterns are observed in InternVL3.5-38B and Gemini~2.5~Pro, whose large gains from strict to relaxed accuracy indicate that errors mainly occur within close spatial proximity to the target cell. 
These results imply that the current MLLMs have insufficient fine-grained spatial calibration.

\subsection{Effectiveness of OddGrid-GRPO}

As shown in Table~\ref{table_method}, OddGrid-GRPO substantially outperforms both the supervised baseline, standard GRPO, and GSPO~\cite{zheng2025group} across all types of visual discrepancy tasks. 
The Base model struggles to perceive subtle attribute differences, achieving only 17.14\% overall accuracy. 
Applying GRPO markedly enhances fine-grained discrimination to 70.86\%, verifying the effectiveness of RL optimization. 
Building upon this, OddGrid-GRPO further raises performance to 82.64\%, achieving an additional gain of +11.78\% over GRPO. 
This improvement is consistent across all categories, with particularly notable gains in Rotation (+13.0) and Position (+23.0). 
These results indicate that the proposed distance-aware reward provides continuous feedback proportional to spatial deviation, enabling the model to capture fine-grained discrepancies more effectively rather than relying solely on binary correctness signals. 

In addition to the quantitative gains, the training dynamics shown in Figure~\ref{fig:reward_curve} further validate the effectiveness of OddGrid-GRPO. 
Compared with GRPO, our method converges faster and achieves higher final rewards throughout training, demonstrating enhanced optimization stability and efficiency brought by the distance-aware reward and curriculum-guided schedule.
Ablation experiments further confirm the contribution of each module: removing the distance reward (w/o~$r_{\text{d}}$) leads to a sharp drop to 72.50\%, while disabling curriculum-guided scheduling (w/o~Cur-Guided) reduces accuracy to 78.43\%. 
Together, these results show that distance-sensitive rewards and progressive curricula jointly improve perceptual alignment and training stability, leading to enhanced performance.

\vspace{-1mm}
\section{Conclusion}
\label{sec:conclusion}
In this work, we introduce OddGridBench, a controllable benchmark for evaluating the visual discrepancy sensitivity of MLLMs. 
Our study reveals that fine-grained perceptual sensitivity remains a fundamental bottleneck for current MLLMs, hindering their ability to achieve reliable, grounded visual understanding.  
We further develop OddGrid-GRPO, an RL framework that embeds spatial proximity into the training objective and integrates curriculum-guided optimization with distance-aware rewards to improve training stability and human-aligned visual sensitivity.
We believe OddGridBench and OddGrid-GRPO establish a principled framework for perception-grounded learning, offering new insights into the perceptual foundations of multimodal intelligence.

{
    \small
    \bibliographystyle{ieeenat_fullname}
    \bibliography{main}
}

\clearpage
\setcounter{page}{1}
\maketitlesupplementary
\appendix

\section{Appendix Outline}
In these supplementary materials, we provide:
\begin{itemize}
\item A detailed description of the evaluation setup and full experimental results for all OddGridBench sub-tasks (Appendix~\ref{exp});
\item Additional cross-dataset and cross-format generalization experiments are presented in Appendix~\ref{additional_experiments}.
\item Additional visualizations and qualitative examples (Appendix~\ref{visual}).
\end{itemize}

\section{Experiment Details}
\label{exp}

\subsection{Model Access}
This section summarizes the model access settings and parameter configurations used in our experiments (see Table~\ref{model_access}). All results reported in this paper are based on model outputs obtained prior to November 1, 2025.
\begin{table}[h]
\centering
    \renewcommand\arraystretch{1}
      \caption{List of MLLMs evaluated in this study, with model names shown exactly as they appear on Hugging Face or in official APIs.}
    \label{model_access}
    \large
\scalebox{0.6}{
\begin{tabular}{ll}
\toprule[1pt]
Phi-3.5-vision      & microsoft/Phi-3.5-vision-instruct        \\
SAIL-VL2-2B       & BytedanceDouyinContent/SAIL-VL2-2B                 \\
SAIL-VL2-8B      & BytedanceDouyinContent/SAIL-VL2-8B                \\
LLaVA-OneVision-1.5-4B  & lmms-lab/LLaVA-OneVision-1.5-4B-Instruct \\
LLaVA-OneVision-1.5-8B  & lmms-lab/LLaVA-OneVision-1.5-8B-Instruct \\
LLaVA-v1.6-34B      & liuhaotian/llava-v1.6-34b                \\
InternVL3.5-38B      & OpenGVLab/InternVL3\_5-38B                \\
Molmo-72B      & allenai/Molmo-72B-0924                \\
Qwen2.5-VL-7B       & Qwen/Qwen2.5-VL-7B-Instruct              \\
Qwen2.5-VL-72B      & Qwen/Qwen2.5-VL-72B-Instruct             \\ 
Qwen3-VL-2B       & Qwen/Qwen3-VL-2B-Instruct              \\
Qwen3-VL-4B       & Qwen/Qwen3-VL-4B-Instruct              \\
Qwen3-VL-8B       & Qwen/Qwen3-VL-8B-Instruct              \\
Qwen3-VL-30B (Moe)      & Qwen/Qwen3-VL-30B-A3B-Instruct              \\ 
Qwen3-VL-32B       & Qwen/Qwen3-VL-32B-Instruct              \\ \midrule
Gemini-2.0-Flash    & gemini-2.0-flash                        \\
Gemini-2.5-Flash    & gemini-2.5-flash                     \\
Gemini-2.5-Pro    & gemini-2.5-pro                      \\
GPT-5      & gpt-5-low                      \\ \bottomrule[1pt]
\end{tabular}}
\end{table}

\subsection{Evaluation Details}
\paragraph{Evaluation Prompt.} We use a fixed prompt that instructs the model to identify the odd element in the grid and return its position in a standardized format. We set \textit{max\_new\_tokens}=1024 and use default values for all other parameters.

\begin{tcolorbox}[
    colback=gray!5,
    colframe=black!60,
    title=Evaluation Prompt,
    fonttitle=\bfseries,
    width=0.5\textwidth,
    boxrule=0.5pt,
    arc=2pt
]

\textbf{You are solving an odd-one-out visual perception task.} \\
You are given an image showing a \texttt{\{rows\}×\{cols\}} grid of \texttt{\{shape\}}s.  
All \texttt{\{shape\}}s appear the same, except one that is visually different in \texttt{\{odd\_desc\}}.

\vspace{2mm}

This is a \textbf{visual perception} task that does not require lengthy logical reasoning.

\vspace{2mm}
\textbf{Instructions}
\begin{itemize}
    \item Carefully inspect the grid.
    \item Identify the grid position (row and column) of the \texttt{\{shape\}} that is different.
    \item Counting starts from the top-left corner, i.e., Row~1, Column~1.
    \item Provide brief visual observations if needed (no more than 300 words).
\end{itemize}

\textbf{Output Format Requirements}
\begin{itemize}
    \item Provide concise natural-language observations.
    \item End the response with the final answer in the following strict LaTeX format:
\begin{verbatim}
\boxed{Row X, Column Y}
\end{verbatim}
    where \texttt{X} and \texttt{Y} are integers (e.g., \texttt{Row 2, Column 3}).
    \item Do \textbf{not} include any text after the final \verb|\boxed{}|.
    \item If no odd \texttt{\{shape\}} exists, output:
\begin{verbatim}
\boxed{Row 0, Column 0}
\end{verbatim}
\end{itemize}

\end{tcolorbox}

\paragraph{Human Evaluation Protocol.}
To estimate human-level performance on OddGridBench, we conduct a human evaluation on a subset of the dataset by randomly sampling 350 instances (50 from each perceptual attribute category). Three human annotators with backgrounds in computer science and visual perception participate in the evaluation, all with normal or corrected-to-normal vision. The evaluation is conducted using a custom annotation interface implemented in Python. Images are displayed on a 4K-resolution monitor and rendered at a fixed resolution, with zooming or resizing disabled to ensure consistent perceptual conditions. Before the formal evaluation, annotators are provided with several practice examples to familiarize themselves with the task. For each image, participants are asked to identify the single grid element that differs from the others using only visual cues, without any external tools. During the evaluation, images are presented one at a time, and annotators select the position of the discrepant element, which may differ in color, size, rotation, or position. No feedback about correctness is provided during the evaluation.

\subsection{Training Details}
\paragraph{Training Settings.}
We adopt Qwen3-VL-2B-Instruct as the base model and train it using the {EasyR1} reinforcement learning framework.
Training is conducted with a global actor batch size of 256, micro-batch sizes of 1 (for updates) and 2 (for experience logging), and dynamic padding enabled to improve efficiency.
Gradient checkpointing is enabled, and the vision tower remains fully trainable.
For memory management, we use partial FSDP with rank-0 initialization enabled and no CPU offloading.
During rollout, we generate $n{=}3$ samples per prompt using {vLLM} with temperature~1.0, top-$p{=}1.0$, and tensor parallel size~2.
Validation rollouts use a lower temperature (0.6) and narrower sampling (top-$p{=}0.95$).
The reference model is kept frozen and trained under FSDP-full-shard mode.
The reward is computed in batch mode using a task-specific scoring function.
Training runs for 100 optimization steps on a single node with 4~A800 GPUs.
We evaluate every 10 steps, log up to 3 generated samples per validation round, and save checkpoints every 10 steps with a maximum retention of 4. 
For the GSPO variant, we additionally enable token-level GSPO loss, sequence-level advantage averaging, a tighter clipping ratio (\texttt{3e{-}4} to \texttt{4e{-}4}), and disable KL regularization, following the GSPO formulation.

\paragraph{Training Prompt.}
We employ a minimal instruction prompt during training to prevent models from exploiting unnecessary reasoning cues and to ensure that learning focuses purely on visual discrepancy detection.
\begin{tcolorbox}[
    colback=gray!5,
    colframe=black!60,
    title=Training Prompt,
    fonttitle=\bfseries,
    width=0.47\textwidth,
    boxrule=0.5pt,
    arc=2pt
]
Identify the object that differs from the others in the \{rows\}×\{cols\} grid.  
The difference lies in \{odd\_desc\}.  
Count positions from the top-left corner as Row~1, Column~1.  
Return the final answer strictly in the following format:
\begin{verbatim}
\boxed{Row X, Column Y}
\end{verbatim}
\end{tcolorbox}

\subsection{Additional Ablation Results}

\begin{figure}[h]
\centering
\includegraphics[width=\linewidth]{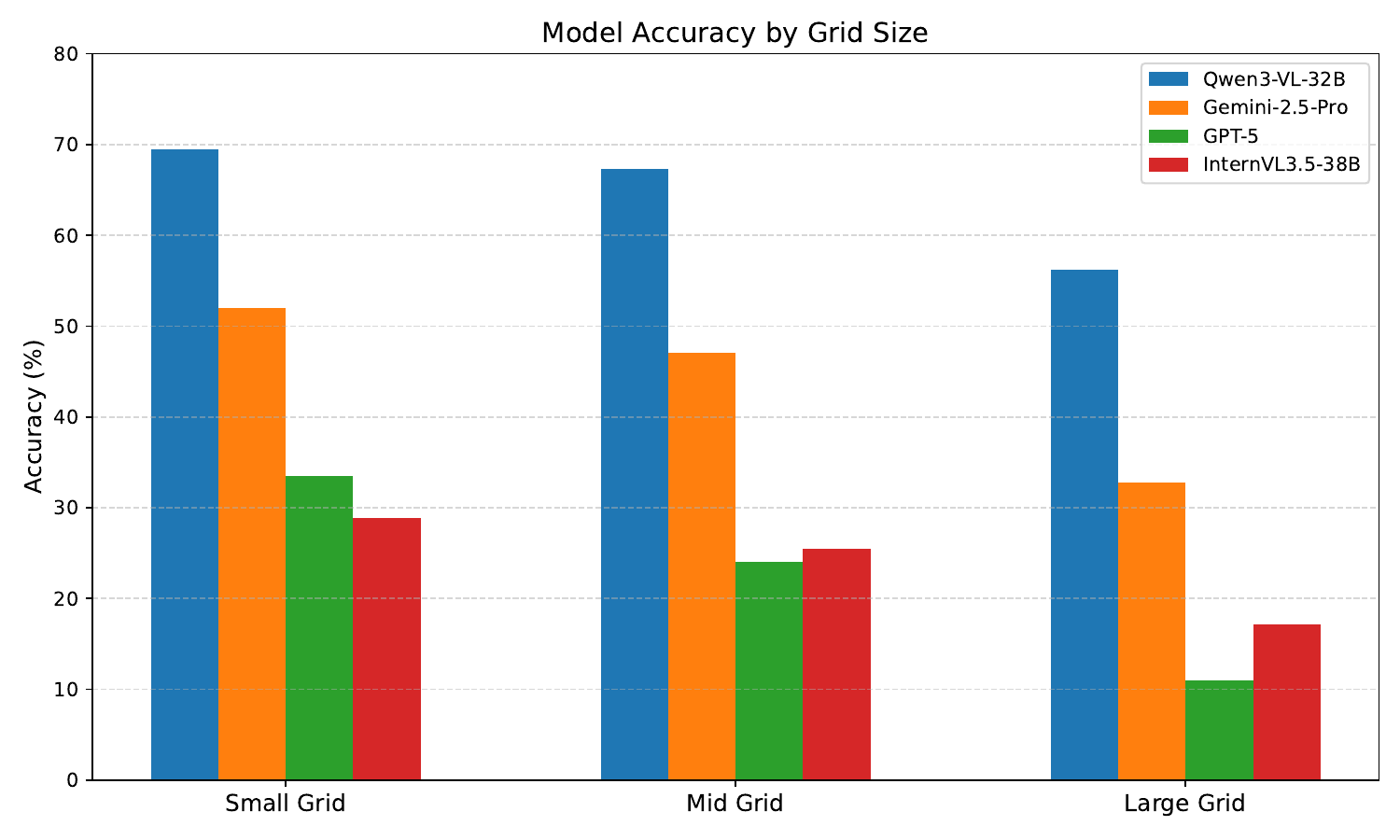}
\caption{
Comparison of MLLMs' performance under different grid densities.
Grid size is categorized by total cell count ($r \times c$): 
{Small Grid} ($\leq 44$ cells), 
{Medium Grid} (45–64 cells), 
and {Large Grid} (65–81 cells).
}
\label{fig:bar}
\end{figure}

\noindent \textbf{Impact of Grid Size on Visual Discrimination.}
To investigate how spatial density affects model perception, we compare MLLM accuracy across three grid configurations: small, medium, and large grids.
As shown in Figure~\ref{fig:bar} and Table~\ref{tab:grid_size}, all models experience a consistent decline in accuracy as the grid size increases.
Larger grids introduce more distractors and visual clutter, which reduces the relative salience of the odd item and makes fine-grained discrimination substantially harder.
Among the evaluated models, Qwen3-VL-4B and Gemini~2.5~Pro maintain moderate robustness under mid-scale grids, while GPT-5 and InternVL~3.5-38B show steeper performance degradation.
This pattern indicates that current MLLMs have limited ability to maintain spatial attention and object separation when visual scenes become more crowded.
\begin{table}[h]
\centering
\caption{Performance across different grid sizes.}
\label{tab:grid_size}
\begin{tabular}{lccc}
\toprule
Model & Grid-Small & Grid-Mid & Grid-Large \\
\midrule
InternVL3.5-38B & 27.74 & 26.94 & 25.42 \\
Qwen3-VL-32B    & 70.49 & 66.97 & 64.41 \\
Gemini-2.5-Pro           & 54.06 & 47.34 & 41.24 \\
GPT-5                    & 35.51 & 26.79 & 15.82 \\
\bottomrule
\end{tabular}
\end{table}

\noindent \textbf{Impact of Image Resolution on Visual Discrimination.} We further conducted an ablation study on different per-cell resolutions. As shown in Table~\ref{tab:imgres}, Gemini-2.5-Pro: accuracy increases from 47.0 at resolution 50 to 57.0 at 100, and then plateaus at 57.0 at 150, while GPT-5 rises from 31.0 to 35.0 to 39.0. Overall, higher resolution yields only modest gains.

\begin{table}[h]
\centering
\caption{Performance across different image resolutions.}
    \label{tab:imgres}
    \begin{tabular}{@{}l ccc@{}} 
        \toprule
        & \multicolumn{3}{c}{{Resolution}} \\
        \cmidrule(l){2-4}
        \textbf{Model} & {50} & {100} & {150} \\
        \midrule
        Qwen3-VL-32B   & 69.00 & 70.00 & 73.00 \\
        Qwen3-VL-8B    & 56.00 & 60.00 & 59.00 \\
        Gemini-2.5-Pro & 47.00 & 57.00 & 57.00 \\
        GPT-5          & 31.00  & 35.00 & 39.00 \\
        \bottomrule
    \end{tabular}
\end{table}

\noindent \textbf{Sensitivity to Perceptual Discrepancy Magnitude.}  
Tables~\ref{tab:qwen3vl}–\ref{tab:intervl} summarize model performance across increasing 
levels of perceptual discrepancy, illustrating how accuracy changes as the visual difference becomes more pronounced.

\begin{figure*}[h]
\centering
\includegraphics[width=\linewidth]{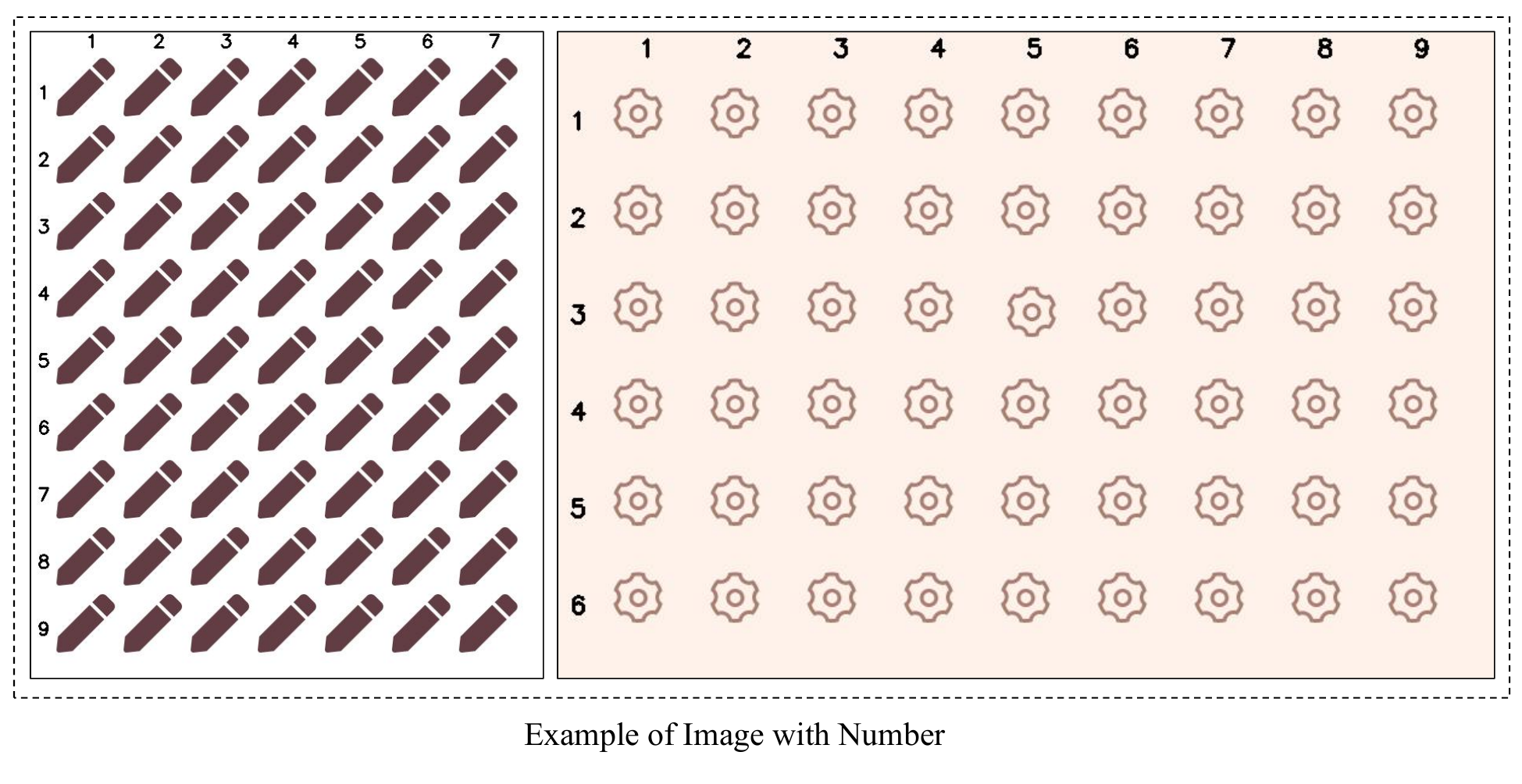}
\caption{
Example of grid images with row and column indices used in {OddGridBench}.
}
\label{fig:number_image}
\end{figure*}

\noindent \textbf{Analysis of Localization Accuracy.}  
Figure~\ref{fig:number_image} illustrates the grid images annotated with row and 
column indices that are used to evaluate models’ localization performance (LabeledAcc).

\noindent \textbf{Impact of Image Resolution}.
In a resolution ablation (50/100/150), Gemini-2.5-Pro improves 47.0→57.0 and then saturates, while GPT-5 increases 31.0→35.0→39.0, indicating only modest gains from higher resolution, as shown in Table~\ref{tab:res_ablation}.

\begin{table}[h]
\centering
\scriptsize
\setlength{\tabcolsep}{3.5pt}
\renewcommand{\arraystretch}{0.5}
\caption{Ablation on image resolution. We evaluate the performance of different models under three input resolutions (50, 100, and 150). Higher resolutions generally provide richer visual details and lead to improved odd-item detection accuracy.}
\label{tab:res_ablation}
\resizebox{0.8\linewidth}{!}{
\begin{tabular}{@{}l ccc@{}}
\toprule
& \multicolumn{3}{c}{Resolution} \\
\cmidrule(l){2-4}
\textbf{Model} & 50 & 100 & 150 \\
\midrule
Gemini-2.5-Pro & 47.00 & 57.00 & 57.00 \\
GPT-5          & 31.00 & 35.00 & 39.00 \\
\bottomrule
\end{tabular}
}
\vspace{-3mm}
\end{table}

\noindent \textbf{Why Row–Column Instead of Bounding-Box Grounding?}
Although bounding-box localization appears to be a more fine-grained grounding objective, we intentionally adopt the \texttt{Row X, Column Y} answer format in OddGridBench for two reasons.
{First}, the OddGrid setting provides a discretized grid structure in which the notion of “location” is inherently symbolic rather than continuous; requiring pixel-level bounding boxes would not introduce additional reasoning difficulty, but rather introduce annotation ambiguity (e.g., bounding box tightness, padding, icon margins) that is irrelevant to the core perceptual challenge.
{Second}, as shown in Table~\ref{tab:iou_qwen}, even strong MLLMs that correctly predict the grid position fail almost completely when evaluated with IoU-based box matching, confirming that low-level spatial grounding remains a confounding variable that masks the actual perceptual reasoning ability we aim to measure.
For these reasons, we adopt the row–column formulation to ensure that OddGridBench faithfully evaluates fine-grained visual discrimination rather than geometric box regression.

\noindent \textbf{Error Anylisis.}
We treat cases where the predicted location is close to the ground truth as a proxy for the linguistic mapping challenge (23.0\% for Qwen3-VL-32B; 20.9\% for Gemini-2.5-Pro).
In contrast, large-deviation failures dominate (77.0\% and 79.1\%), suggesting the main bottleneck is visual inability. 

\begin{tcolorbox}[
    colback=gray!5,
    colframe=black!60,
    title=Bounding Box Prompt,
    fonttitle=\bfseries,
    width=0.47\textwidth,
    boxrule=0.5pt,
    arc=2pt
]
Several small grids are shown in the image, but you should focus on the global image coordinates.  
First, briefly describe what you observe in the image and how one object differs from the others.  
Then, identify the object that differs from all others in the entire image.  
Return its bounding box in the \textbf{global image coordinate system} strictly in the following format:
\begin{verbatim}
\boxed{[x_min, y_min, x_max, y_max]}
\end{verbatim}
\end{tcolorbox}

\begin{table}[h]
\centering
\renewcommand{\arraystretch}{0.9}
\caption{Bounding-box localization accuracy of Qwen3-VL models at different IoU thresholds.}
\begin{tabular}{lccc}
\toprule
\textbf{Model} & \textbf{IoU@0.3} & \textbf{IoU@0.4} & \textbf{IoU@0.5} \\
\midrule
Qwen3-VL-2B  & 0.074 & 0.053 & 0.036 \\
Qwen3-VL-4B  & 0.089 & 0.062 & 0.041 \\
Qwen3-VL-8B  & 0.102 & 0.078 & 0.055 \\
Qwen3-VL-32B & 0.133 & 0.087 & 0.053 \\
\bottomrule
\end{tabular}
\label{tab:iou_qwen}
\end{table}

\section{Additional Cross-Dataset and Cross-Format Experiments}
\label{additional_experiments}

\subsection{Cross-Dataset Generalization}

\begin{figure}[h]
\centering
\includegraphics[width=\linewidth]{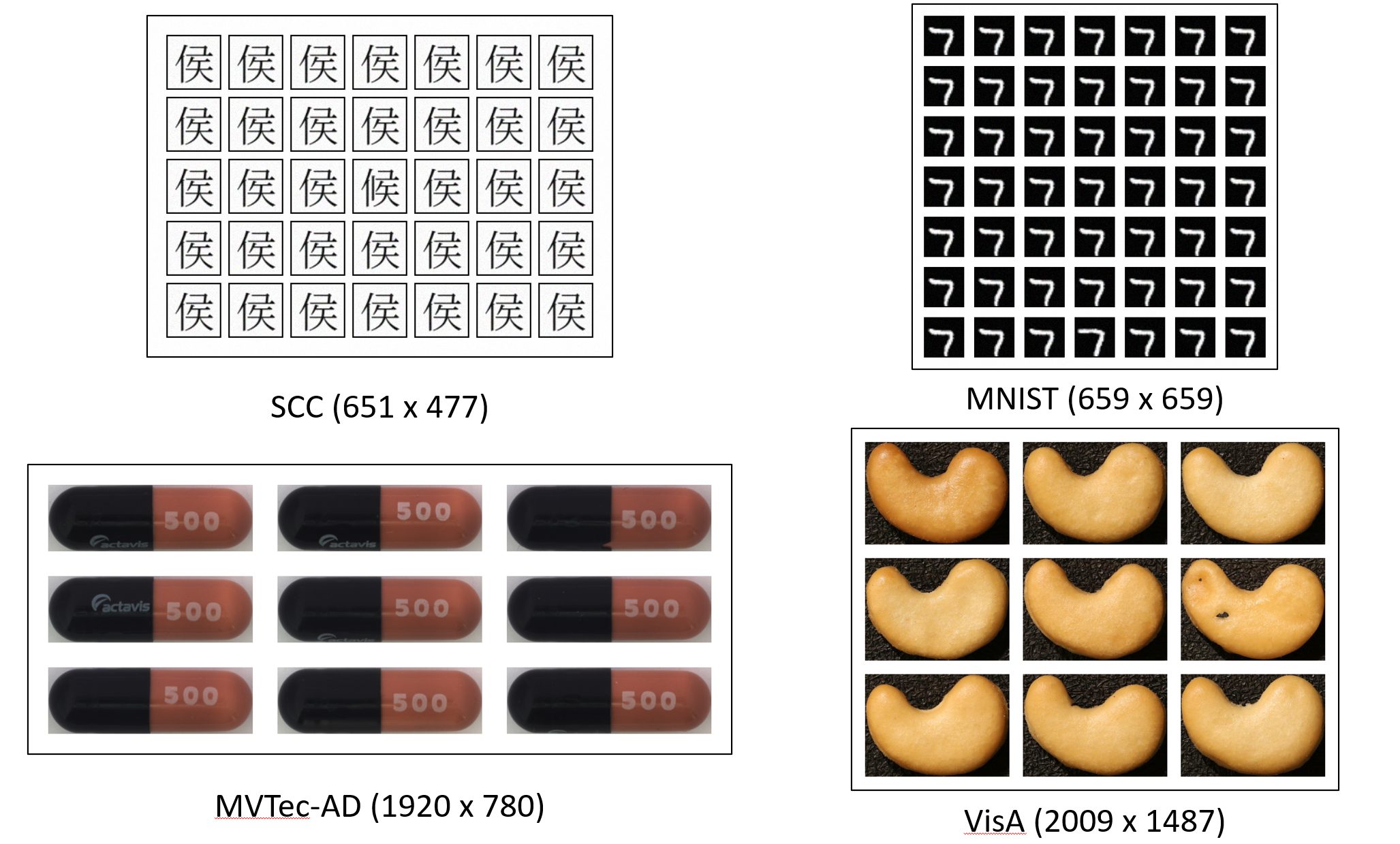}
\caption{
Visualization of the four datasets used in the image sequence settings.
}
\label{fig:iol}
\end{figure}

We conducted extended experiments on four test sets.
\textcolor{red}{(1) MVTec-AD and VisA (real-image industrial anomaly detection).}
We use 13 categories from MVTec-AD and 7 from VisA, where defect images (per official ground truth) are treated as odd items and corresponding normal images as distractors.
We crop background regions, keep the longest side of each cell $\geq 500$ px, and organize samples into $3\times3$ to $5\times5$ grids.
\textcolor{red}{(2) MNIST and Similar Chinese Characters (SCC).}
For MNIST and CC, we first select structurally similar instances as base cells (e.g., same digit with cosine similarity $\geq 80\%$ for MNIST; visually similar pairs for SCC), as shown in Figure~\ref{fig:iol}.
The remaining data construction procedure follows the same protocol as OddGridBench. Each dataset contains 100 grid samples, resulting in a total of 400 evaluation instances across four datasets.

We evaluate the checkpoints on these test sets, as shown in Table~\ref{tab:cross-data}. Our method consistently outperforms both the baseline and standard GRPO across all datasets, indicating transfer beyond synthetic icons to real-image settings. 
We will include these extensions and further analysis in the revised version.

\begin{table}[!t]
\centering
\renewcommand{\arraystretch}{1.0}
\caption{Cross-dataset generalization results on four external datasets, including two real-image anomaly detection benchmarks (MVTec-AD and VisA) and two handwritten/character datasets (MNIST and SCC).}
\label{tab:cross-data}
\resizebox{0.95\linewidth}{!}{
\begin{tabular}{lcccc}
\toprule
Method & MVTec-AD & VisA & MNIST & SCC \\ \midrule
Qwen3-VL-32B & 67.00 & 66.00 & 39.00 & 87.00 \\ \midrule
Qwen3-VL-2B (Baseline) & 20.00 & 9.00 & 6.00 & 17.00 \\ 
GRPO & {47.00} & 39.00 & 18.00 & 54.00 \\
\textbf{OddGrid-GRPO (Ours)} & \textbf{49.00} & \textbf{40.00} & \textbf{37.00} & \textbf{60.00} \\ 
\bottomrule
\end{tabular}
}
\end{table}

\vspace{-2mm}
\subsection{Cross-Format Generalization}
\begin{figure}[h]
\centering
\includegraphics[width=\linewidth]{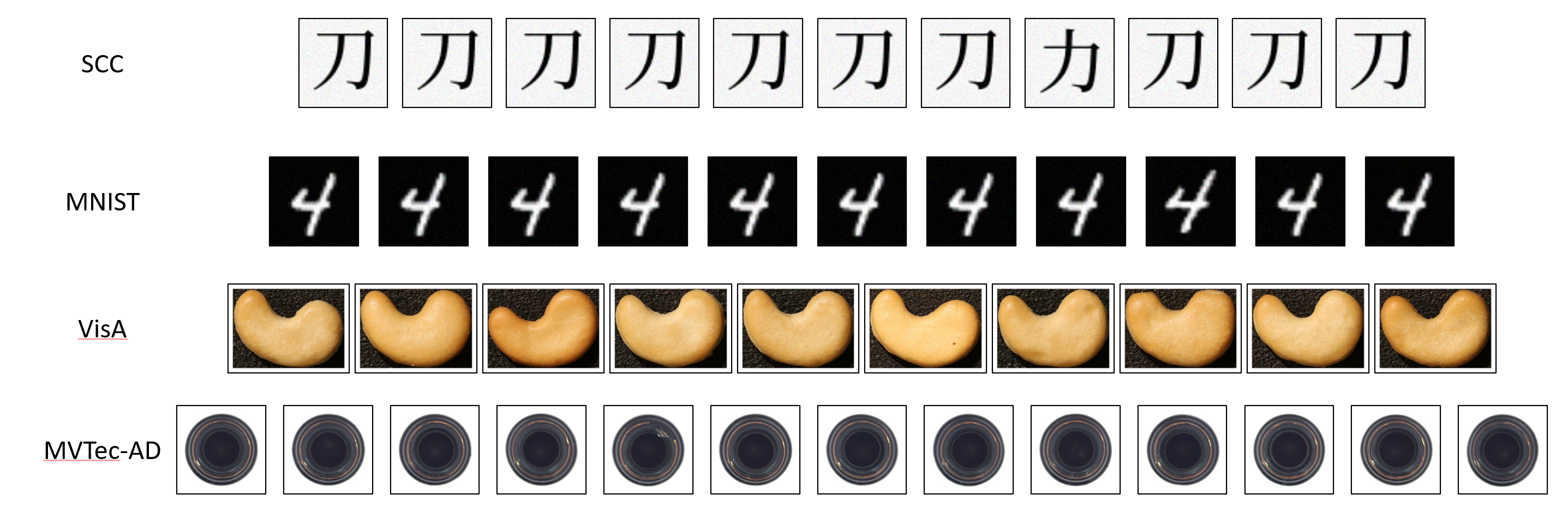}
\caption{
Visualization of the four datasets used in the row and column settings.
}
\label{fig:soi}
\end{figure}

We evaluate a non-grid variant (as shown in Figure~\ref{fig:soi}) where the model receives 8--15 separate images and predicts the odd-one-out index, Table~\ref{tab:cross-format}.
OddGrid-GRPO outperforms both the baseline and standard GRPO, including on real-image anomaly detection datasets (MVTec-AD/VisA), suggesting gains beyond grid-specific memorization. We will add qualitative examples in the revised version.

\begin{table}[!t]
\centering
\renewcommand{\arraystretch}{1.1}
\caption{Cross-format generalization results under a non-grid setting where the model receives 8--15 independent images and predicts the odd-one-out index.}
\label{tab:cross-format}
\resizebox{0.95\linewidth}{!}{
\begin{tabular}{lcccc}
\toprule
Method & MVTec-AD & VisA & MNIST & SCC \\ \midrule
Qwen3-VL-32B & 76.00 & 72.00 & 62.00 & 86.00 \\ \midrule
Qwen3-VL-2B (Baseline) & 29.00 & 32.00 & 2.00 & 28.00 \\ 
GRPO & {42.00} & 38.00 & 11.00 & 51.00 \\
\textbf{OddGrid-GRPO (Ours)} & \textbf{46.00} & \textbf{40.00} & \textbf{16.00} & \textbf{64.00} \\ 
\bottomrule
\end{tabular}}
\end{table}

\begin{tcolorbox}[
    colback=gray!5,
    colframe=black!60,
    title=Evaluation Prompt (Row and Column type),
    fonttitle=\bfseries,
    width=0.47\textwidth,
    boxrule=0.5pt,
    arc=2pt
]
An image containing a grid of objects is provided.

Identify the object that differs from the others in the $\textit{rows} \times \textit{cols}$ grid.
Cells are indexed starting from the top-left corner as \textbf{Row 1, Column 1}.

Return your answer strictly in the following format:
\begin{verbatim}
\boxed{Row X, Column Y}
\end{verbatim}
\end{tcolorbox}

\begin{tcolorbox}[
    colback=gray!5,
    colframe=black!60,
    title=Image Sequence Identification Prompt,
    fonttitle=\bfseries,
    width=0.47\textwidth,
    boxrule=0.5pt,
    arc=2pt
]

You are presented with \textit{N} images, labeled \texttt{image1}, \texttt{image2}, $\dots$, \texttt{imageN}. Identify all anomalous images in the set.

\textbf{Output Rules:}
\begin{enumerate}
\item You may perform observation and comparative reasoning before answering.
\item The final answer must be enclosed in exactly one \texttt{\textbackslash boxed\{\}} block.
\item Inside the box, list the labels of all anomalous images (e.g., \texttt{image1,image3}).
\item If no anomalous images are found, output \texttt{\textbackslash boxed\{\}}.
\end{enumerate}

\end{tcolorbox}

\subsection{Evaluation on Real-World Anomaly Detection Benchmarks.}

We argue that discrepancy perception is attribute-agnostic, and current models do not yet fully possess this ability.
To better reflect real-world industrial inspection, we redesign the anomaly-count distribution on MVTec-AD and VisA by additionally including all-normal cases and cases with two anomalous items, as shwon in Table~\ref{tab:real-world}. We also revise the prompt to require the model to output the locations of all anomalies. Under EM/F1, even Qwen3-VL-32B reaches only 47.08/58.46 on MVTec-AD and 40.00/53.48 on VisA, indicating that robust real-image discrepancy detection remains challenging.

\begin{table}[h]
\vspace{-2mm}
\centering
\scriptsize
\setlength{\tabcolsep}{3pt}
\caption{Evaluation on real-world industrial anomaly detection datasets. 
We report Exact Match (EM) and F1 scores on MVTec-AD and VisA.}
\label{tab:real-world}
\resizebox{0.8\linewidth}{!}{
\begin{tabular}{lcccc}
\toprule
\multirow{2}{*}{Model} 
& \multicolumn{2}{c}{MVTec-AD} 
& \multicolumn{2}{c}{VisA} \\
\cmidrule(lr){2-3}\cmidrule(lr){4-5}
& EM & F1 & EM & F1 \\
\midrule
Qwen3-VL-8B  & 37.18 & 48.63 & 29.71 & 43.30 \\
Qwen3-VL-32B & 47.08 & 58.46 & 40.00 & 53.48 \\
\bottomrule
\end{tabular}
}
\vspace{-2mm}
\end{table}

\newpage

\begin{table*}[t]
\centering
\renewcommand{\arraystretch}{1.1}
\caption{Accuracy across increasing discrepancy levels on different odd types (Qwen3-VL-32B) 
Higher $\Delta$ levels correspond to larger visual differences.}
\label{tab:qwen3vl}
\begin{tabular}{lccccc}
\toprule
\textbf{Odd Type} & \textbf{$\Delta$ Level 1} & \textbf{$\Delta$ Level 2} & \textbf{$\Delta$ Level 3} & \textbf{$\Delta$ Level 4} & \textbf{$\Delta$ Level 5} \\
\midrule
Color     & 53.66 & 79.07 & 92.31 & 97.96 & 96.55 \\
Position  & 31.25 & 28.57 & 44.26 & 53.85 & 63.64 \\
Rotation  & 38.10 & 31.48 & 67.65 & 70.59 & 69.44 \\
Size      & 7.50  & 25.00 & 50.00 & 52.17 & 67.86 \\
\bottomrule
\end{tabular}
\end{table*}

\begin{table*}[t]
\centering
\renewcommand{\arraystretch}{1.1}
\caption{Accuracy across increasing discrepancy levels on different odd types (Gemini-2.5-Pro). 
Higher $\Delta$ levels correspond to larger visual differences.}
\label{tab:gemini}
\begin{tabular}{lccccc}
\toprule
\textbf{Odd Type} & \textbf{$\Delta$ Level 1} & \textbf{$\Delta$ Level 2} & \textbf{$\Delta$ Level 3} & \textbf{$\Delta$ Level 4} & \textbf{$\Delta$ Level 5} \\
\midrule
Color     & 58.54 & 69.77 & 86.54 & 89.80 & 94.83 \\
Position  & 6.25  & 4.29  & 9.84  & 3.85  & 9.09  \\
Rotation  & 14.29 & 16.67 & 32.35 & 41.18 & 33.33 \\
Size      & 2.50  & 5.00  & 10.87 & 13.04 & 17.86 \\
\bottomrule
\end{tabular}
\end{table*}

\begin{table*}[t]
\centering
\renewcommand{\arraystretch}{1.1}
\caption{Accuracy across increasing discrepancy levels on different odd types (GPT-5). 
Higher $\Delta$ levels correspond to larger visual differences.}
\label{tab:gpt}
\begin{tabular}{lccccc}
\toprule
\textbf{Odd Type} & \textbf{$\Delta$ Level 1} & \textbf{$\Delta$ Level 2} & \textbf{$\Delta$ Level 3} & \textbf{$\Delta$ Level 4} & \textbf{$\Delta$ Level 5} \\
\midrule
Color     & 39.02 & 53.49 & 59.62 & 57.14 & 67.24 \\
Position  & 0.00  & 2.86  & 8.20  & 7.69  & 9.09  \\
Rotation  & 4.76  & 22.22 & 17.65 & 26.47 & 36.11 \\
Size      & 0.00  & 12.50 & 4.35  & 15.22 & 17.86 \\
\bottomrule
\end{tabular}
\end{table*}

\begin{table*}[t]
\centering
\renewcommand{\arraystretch}{1.1}
\caption{Accuracy across increasing discrepancy levels on different odd types (InternVL3.5-38B). 
Higher $\Delta$ levels correspond to larger visual differences.}
\label{tab:intervl}
\begin{tabular}{lccccc}
\toprule
\textbf{Odd Type} & \textbf{$\Delta$ Level 1} & \textbf{$\Delta$ Level 2} & \textbf{$\Delta$ Level 3} & \textbf{$\Delta$ Level 4} & \textbf{$\Delta$ Level 5} \\
\midrule
Color     & 14.63 & 44.19 & 51.92 & 51.02 & 60.34 \\
Position  & 9.38  & 5.71  & 9.84  & 19.23 & 18.18 \\
Rotation  & 0.00  & 11.11 & 14.71 & 32.35 & 27.78 \\
Size      & 0.00  & 5.00  & 2.17  & 0.00  & 3.57  \\
\bottomrule
\end{tabular}
\end{table*}


\section{Example Data and Model Outputs}
\label{visual}
\begin{figure*}[!t]
\centering
\includegraphics[width=\linewidth]{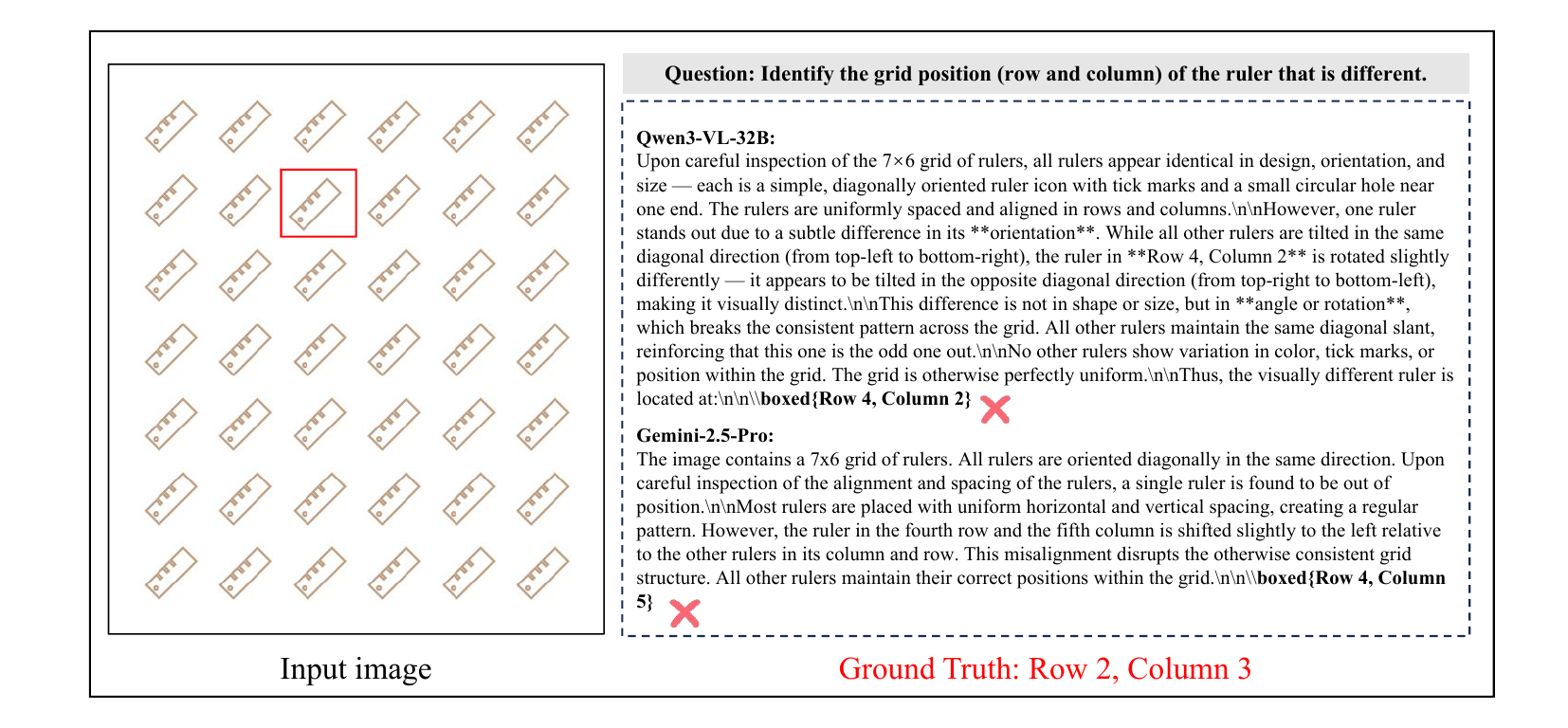}
\caption{
Example of model outputs for fine-grained perceptual discrimination on {OddGridBench}, 
showing predictions from Qwen3-VL-32B and Gemini-2.5 Pro (Example 1/6).}
\label{fig:example_1}
\end{figure*}

\begin{figure*}[!t]
\centering
\includegraphics[width=\linewidth]{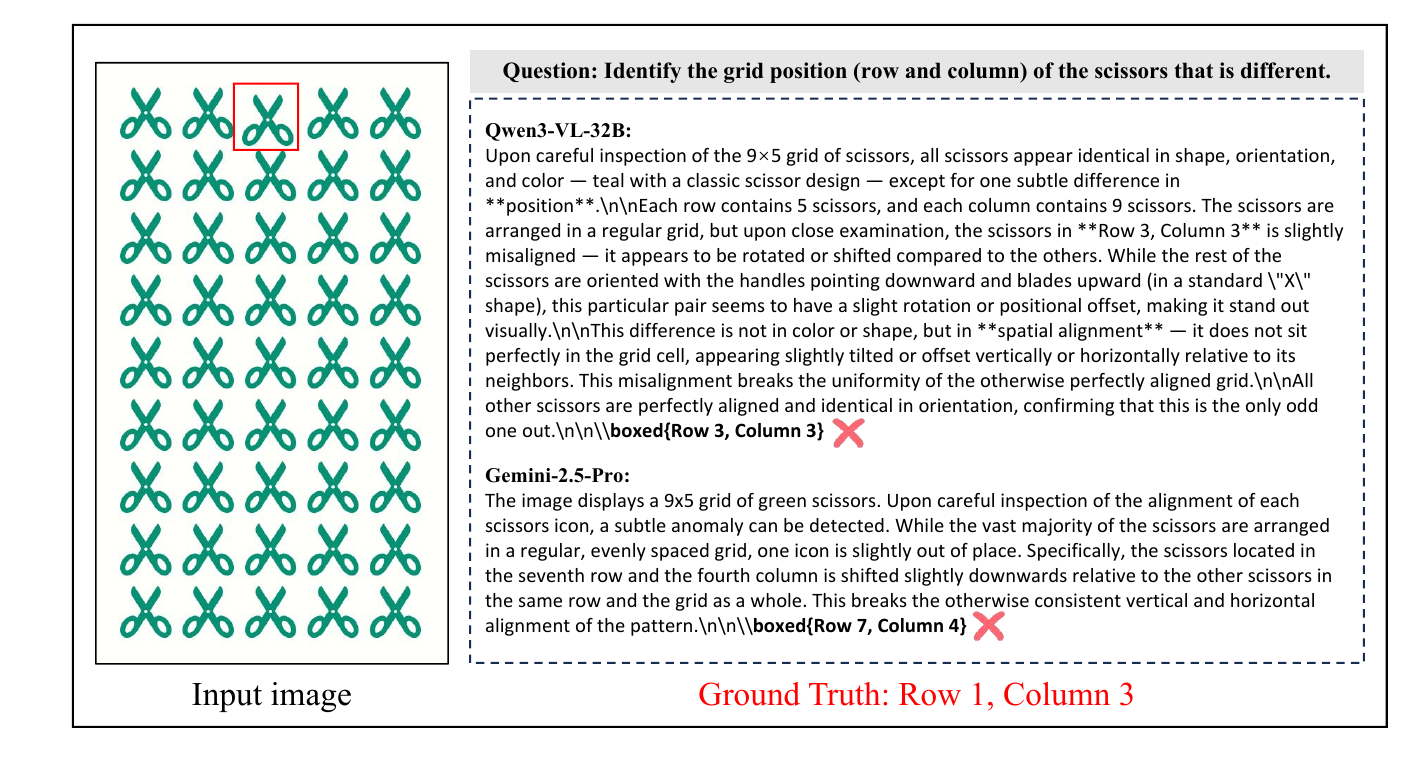}
\caption{
Example of model outputs for fine-grained perceptual discrimination on {OddGridBench}, 
showing predictions from Qwen3-VL-32B and Gemini-2.5 Pro (Example 2/6).}
\label{fig:example_2}
\end{figure*}

\begin{figure*}[!t]
\centering
\includegraphics[width=\linewidth]{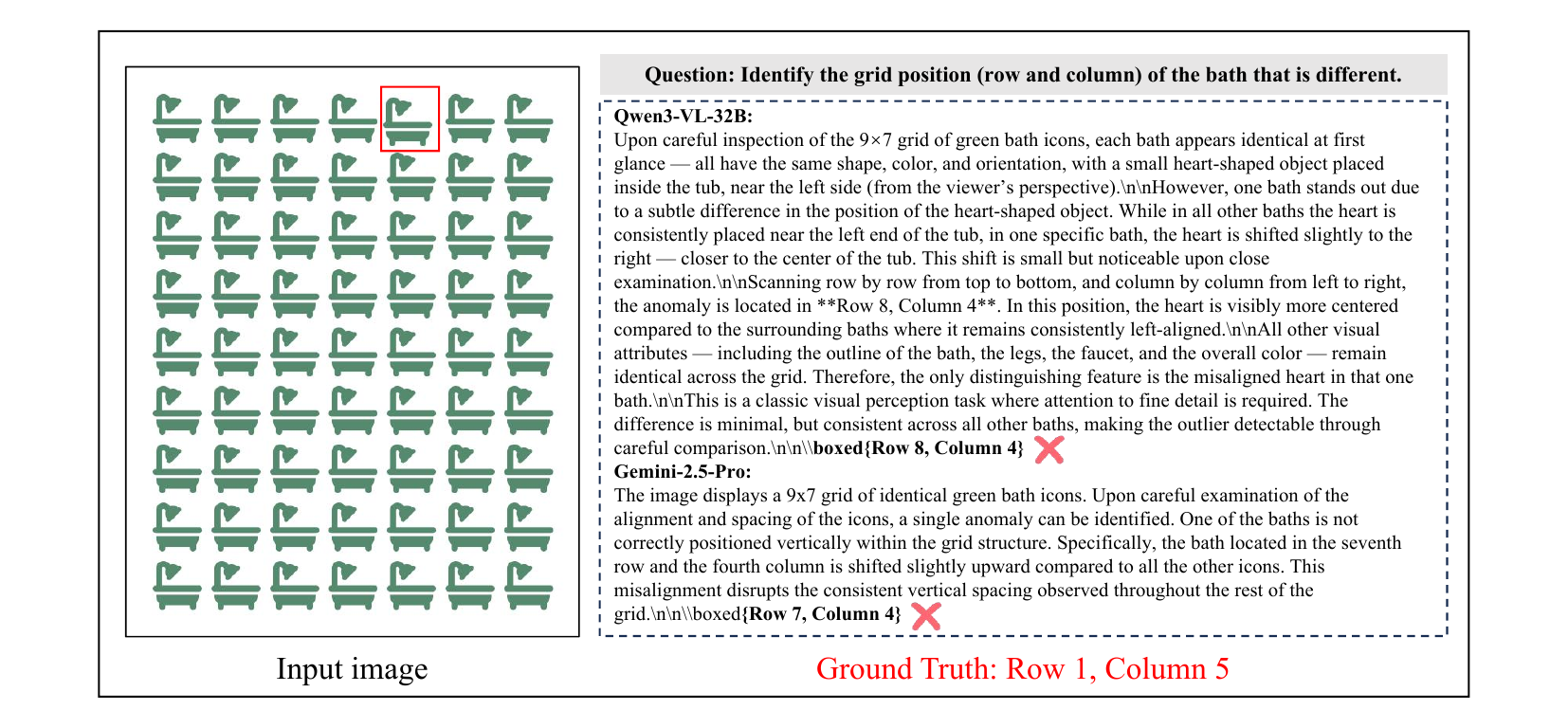}
\caption{
Example of model outputs for fine-grained perceptual discrimination on {OddGridBench}, 
showing predictions from Qwen3-VL-32B and Gemini-2.5 Pro (Example 3/6).}
\label{fig:example_3}
\end{figure*}

\begin{figure*}[!t]
\centering
\includegraphics[width=\linewidth]{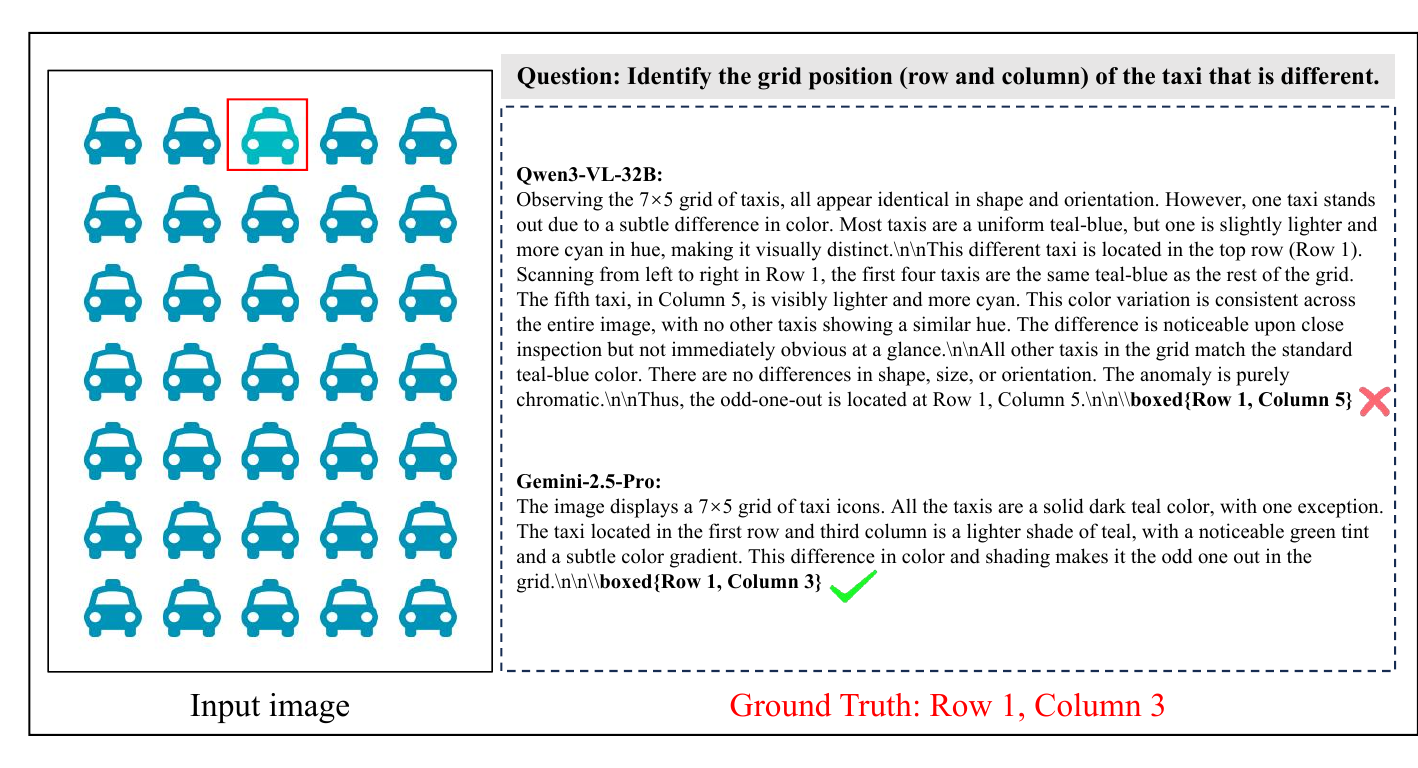}
\caption{
Example of model outputs for fine-grained perceptual discrimination on {OddGridBench}, 
showing predictions from Qwen3-VL-32B and Gemini-2.5 Pro (Example 4/6).}
\label{fig:example_4}
\end{figure*}

\begin{figure*}[!t]
\centering
\includegraphics[width=\linewidth]{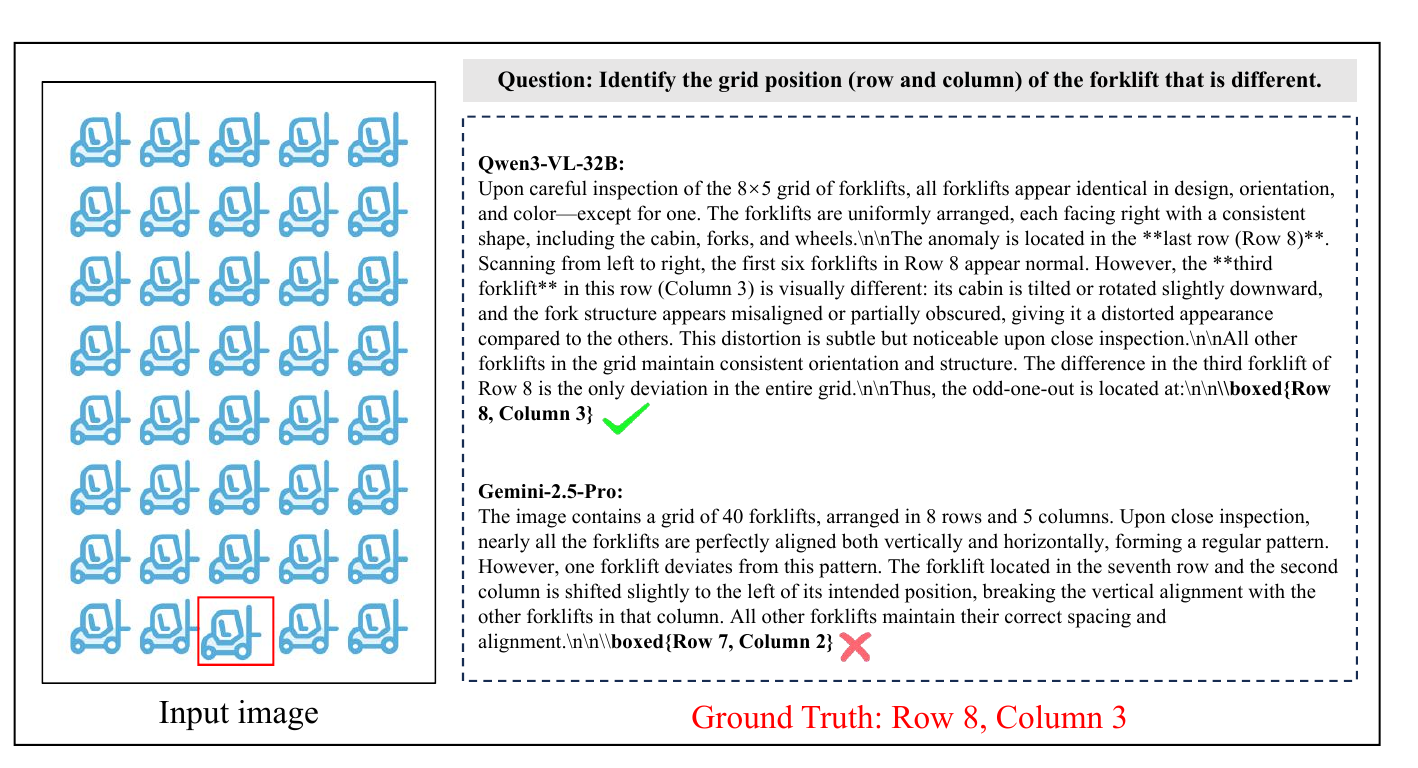}
\caption{
Example of model outputs for fine-grained perceptual discrimination on {OddGridBench}, 
showing predictions from Qwen3-VL-32B and Gemini-2.5 Pro (Example 5/6).}
\label{fig:example_5}
\end{figure*}

\begin{figure*}[!t]
\centering
\includegraphics[width=\linewidth]{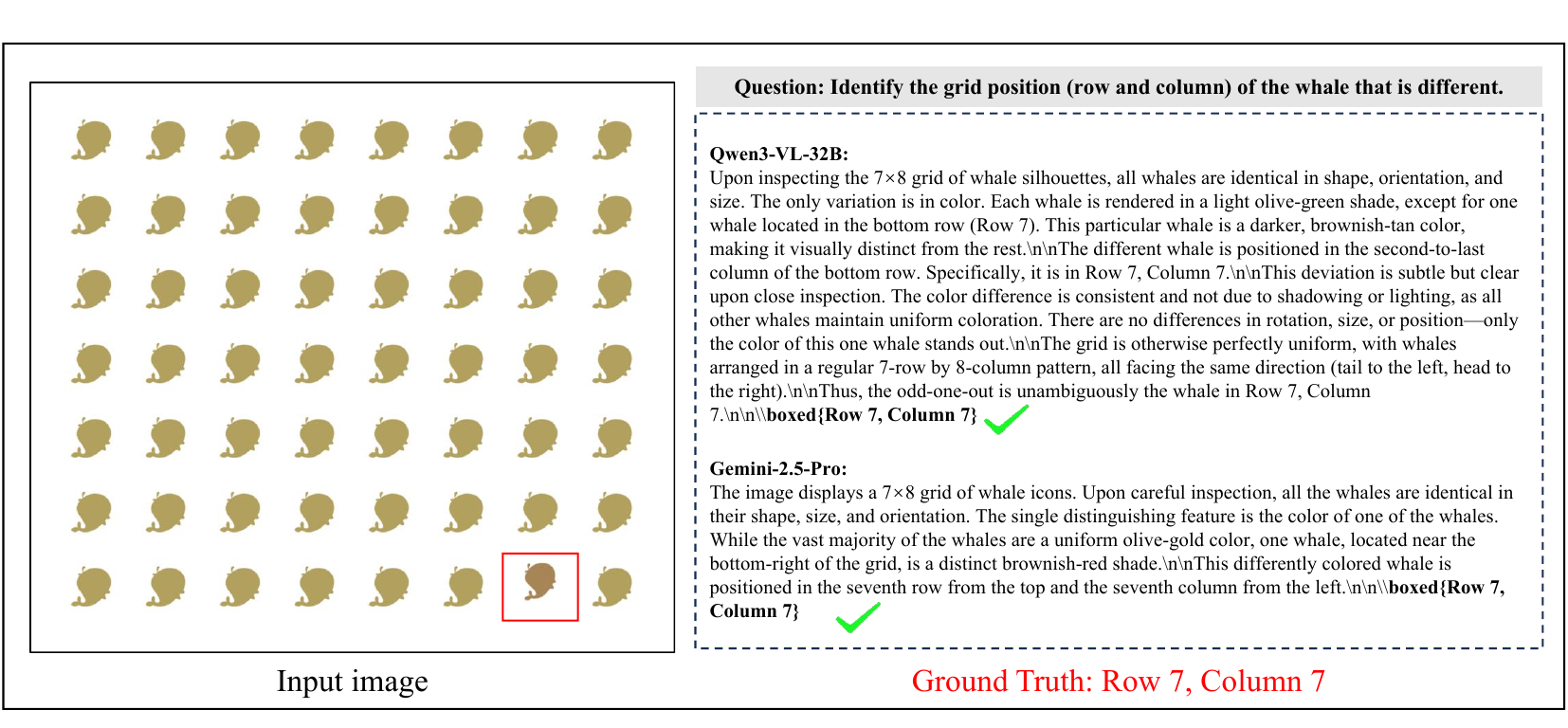}
\caption{
Example of model outputs for fine-grained perceptual discrimination on {OddGridBench}, 
showing predictions from Qwen3-VL-32B and Gemini-2.5 Pro (Example 6/6).}
\label{fig:example_6}
\end{figure*}

\begin{figure*}[!t]
\centering
\includegraphics[width=\linewidth]{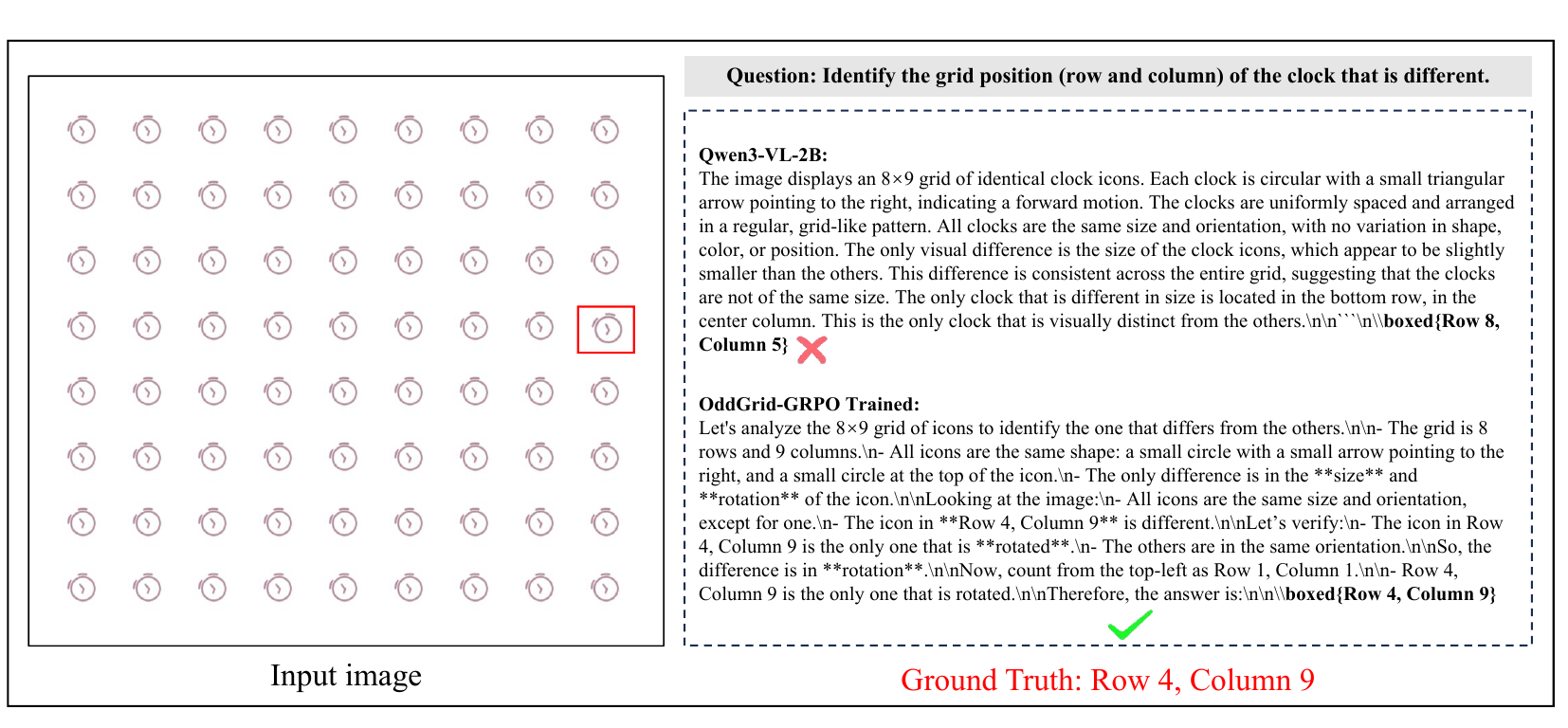}
\caption{
Example of model outputs for fine-grained perceptual discrimination on {OddGridBench}, 
showing predictions from Qwen3-VL-2B and oddGrid-GRPO trained model (Example 1/3).}
\label{fig:example_7}
\end{figure*}

\begin{figure*}[!t]
\centering
\includegraphics[width=\linewidth]{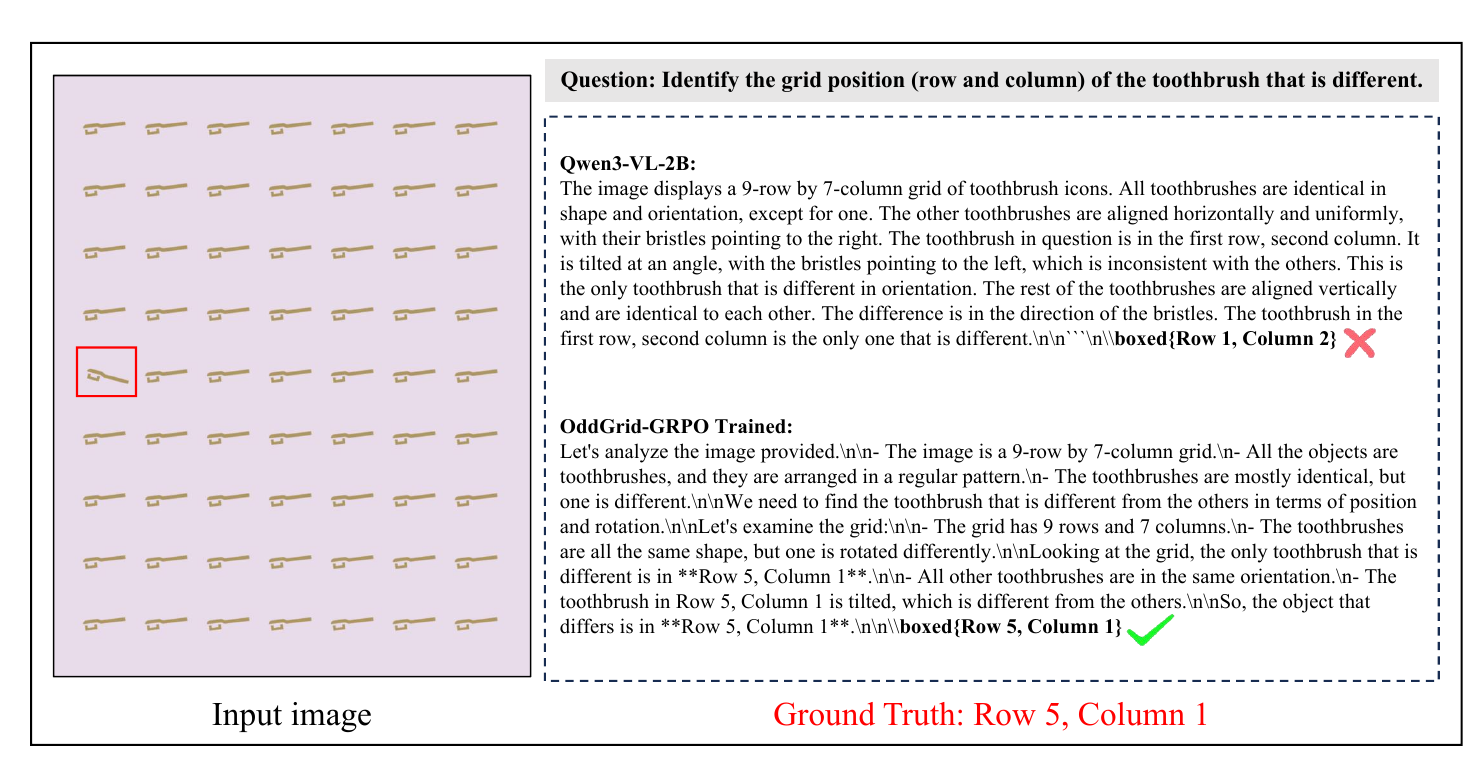}
\caption{
Example of model outputs for fine-grained perceptual discrimination on {OddGridBench}, 
showing predictions from Qwen3-VL-2B and oddGrid-GRPO trained model (Example 2/3).}
\label{fig:example_8}
\end{figure*}

\begin{figure*}[!t]
\centering
\includegraphics[width=\linewidth]{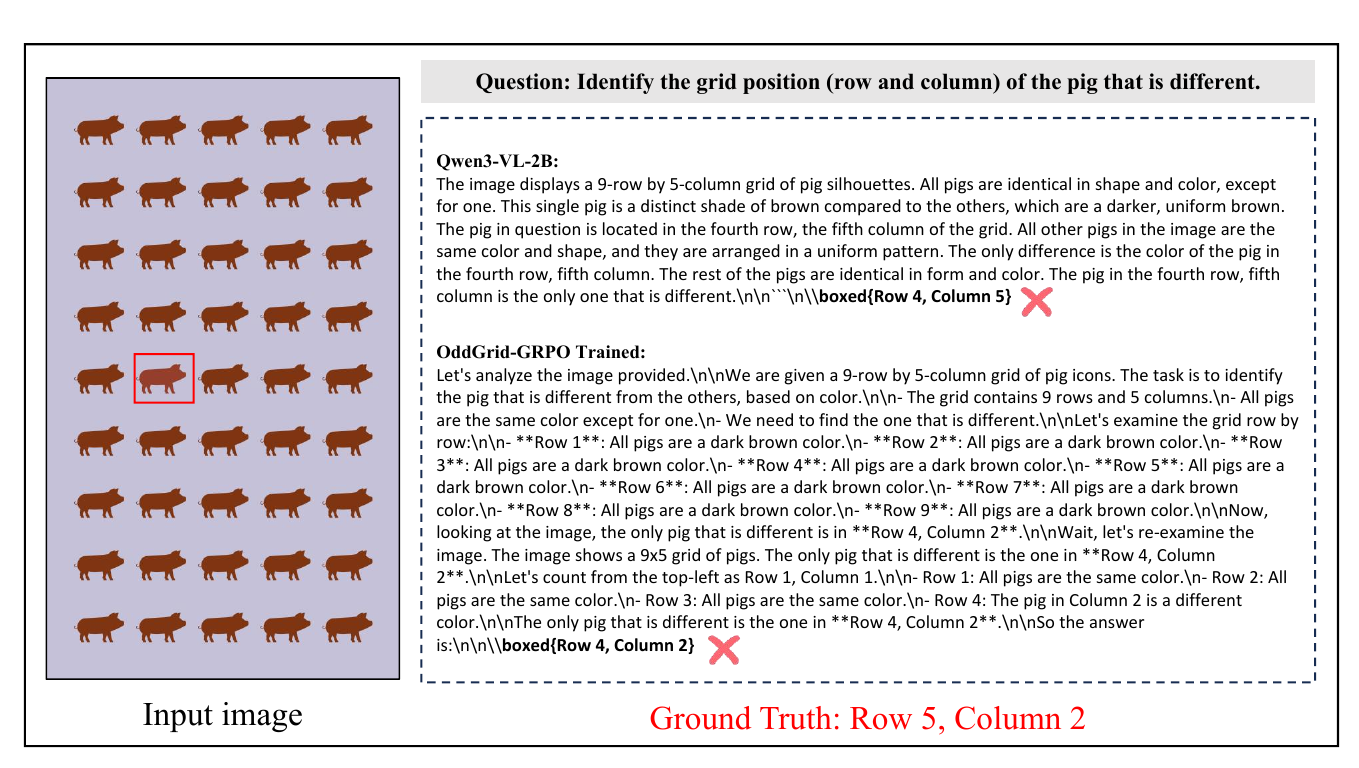}
\caption{
Example of model outputs for fine-grained perceptual discrimination on {OddGridBench}, 
showing predictions from Qwen3-VL-2B and oddGrid-GRPO trained model (Example 3/3).}
\label{fig:example_9}
\end{figure*}

Figures~\ref {fig:example_1} to ~\ref{fig:example_6} show examples from OddGridBench and the responses of Qwen3-VL-32B and Gemini-2.5-Pro.
Figures~\ref {fig:example_7} to ~\ref{fig:example_9} show examples from OddGridBench and the responses of Qwen3-VL-2B and the oddGrid-GRPO trained model.

\end{document}